\documentclass[10pt,twocolumn,letterpaper]{article}

\usepackage[pagenumbers]{cvpr} % To force page numbers, e.g. for an arXiv version

\usepackage{graphicx}
\usepackage{amssymb}
\usepackage{booktabs}
\usepackage{amsmath}
\usepackage[usestackEOL]{stackengine}

\newcommand{\tablestyle}[2]{\setlength{\tabcolsep}{#1}\renewcommand{\arraystretch}{#2}\centering\footnotesize}
\usepackage[pagebackref,breaklinks,colorlinks]{hyperref}

\usepackage[capitalize]{cleveref}
\crefname{section}{Sec.}{Secs.}
\Crefname{section}{Section}{Sections}
\Crefname{table}{Table}{Tables}
\crefname{table}{Tab.}{Tabs.}

\def\cvprPaperID{9158} % *** Enter the CVPR Paper ID here
\def\confName{CVPR}
\def\confYear{2022}

\begin{document}

\title{Routing with Self-Attention for Multimodal Capsule Networks}

\author{%
    Kevin Duarte $^1$ \quad
    Brian Chen $^2$  \quad
    Nina Shvetsova $^3$ \quad
    Andrew Rouditchenko$^4$  \quad
    Samuel Thomas$^{5,6}$  \vspace{1mm} \\
    Alexander Liu$^4$ \quad
    David Harwath$^7$  \quad 
    James Glass$^4$ \quad 
    Hilde Kuehne$^{3,6}$ \quad
    Mubarak Shah$^{1}$ 
    \vspace{1mm} \\
    \small{$^1$University of Central Florida, $^2$Columbia University, $^3$Goethe University Frankfurt,  $^4$MIT CSAIL},  \\
    \small{$^5$IBM Research AI, $^6$MIT-IBM Watson AI Lab, $^7$UT Austin} \\
    \small{
    \texttt{kevin\_duarte@knights.ucf.edu}, \texttt{bc2754@columbia.edu}, \texttt{\{shvetsov,kuehne\}@uni-frankfurt.de} }  \\ \small{\texttt{\{roudi,glass\}@mit.edu}, \texttt{sthomas@us.ibm.com}, 
    \texttt{harwath@cs.utexas.edu}, \texttt{shah@crcv.ucf.edu}   }  
}

\maketitle

\begin{abstract}
% With the growing amount of data, multimodal self-supervised learning has seen a growing interest over the last years.
The task of multimodal learning has seen a growing interest recently as it allows for training neural architectures based on different modalities such as vision, text, and audio. 
%
%Resulting models can be directly applied to practical tasks like text-based retrieval or zero-shot classification. 
%
One challenge in training such models is that
%the fact that, in order to achieve good performance, 
they need to 
%be able to 
jointly learn semantic concepts and their relationships across different input representations.
Capsule networks have been shown to perform well in context of capturing the relation between low-level input features and higher-level concepts. However, capsules have so far mainly been used only in small-scale fully supervised settings due to the resource demand of conventional routing algorithms.
In this work, we propose a new multimodal capsule network that allows us to leverage the strength of capsules in the context of a multimodal learning framework on large amounts of video data.   
To adapt the capsules to large-scale input data, we propose a novel routing by self-attention mechanism that selects relevant capsules which are then used to generate a final joint multimodal feature representation.
This allows not only for robust training with noisy video data, but also to scale up the size of the capsule network compared to traditional routing methods while still being computationally efficient.
We evaluate the proposed architecture by pretraining it on a large-scale multimodal video dataset and applying it on four datasets in two challenging downstream tasks.
Results show that the proposed multimodal capsule network is not only able to improve results compared to other routing techniques, but also achieves competitive performance on the task of multimodal learning.    

\end{abstract}
\section{Introduction}

With the proliferation of video sharing websites and affordable recording devices, the amount of video data available today has dramatically increased. 
Given that hand annotating this continuously growing stream of data is infeasible, recent research has turned to training networks on such large-scale multimodal data without manual annotation \cite{miech2019howto100m,alwassel2019self,miech2020end}.
%But it also shows that the growing amount of data makes training with hand-annotated data more and more infeasible. Therefore, recent research has turned to training networks with such large-scale multimodal data in a self-supervised way \cite{miech2019howto100m,alwassel2019self,miech2020end}. % 
%Well know example for such self-supervised systems are \eg the recently proposed vision-language architectures, which actually leverage images and accompanying text information to pretrain models, that can than be used for a vast amount of different downstream tasks. 
These works make use of the fact that large amounts of data are available across multiple modalities such as vision, text, and audio, especially like in case of videos.
%Those works make use of the fact that, especially in case of video,  large amount of data across multiple modalities such as vision, text, and audio, is available. %
Beyond learning better feature representations by pretraining on large datasets \cite{alwassel2019self}, such networks are able to relate cross-modal inputs based on the similarity of their internal neural representations \cite{miech2020end,alayrac2020self}, which can be applied to zero-shot tasks like classification or text-to-video retrieval. 
%
%These kinds of networks have also been shown to be effectively in zero-shot settings for classification or text-to-video retrieval [need a reference].
%
% according to their representations' similarity, which can be applied to zero-shot tasks like classification or text-to-video retrieval. 
% visual and textual
%
Especially for the latter case, it becomes necessary to capture similar semantic relationships across very different and low level feature representations, as e.g. video features extracted by a ResNet architecture~\cite{he2016deep} have to be related to bag of words representations of sentences \cite{miech2020end}, or even sound representations extracted from audio waveforms \cite{rouditchenko2020avlnet}.
%Especially for the second scenario, it becomes important to capture similar semantic aspects across very different and low level feature representations. E.g. in the existing scenarios video features from a ResNet architecture have to be related to bag of words representations of sentences \cite{miech2020end} ???  or even sound representation \cite{avlnet} \hkc{not 100 perfect yet}.
These relationships can be captured by training a network that takes pairs of modalities as inputs and predicts a similarity score, or by projecting both representations into a joint embedding space. In the second case, for example, the encoding for a sentence like ``Cut the chicken." would be close to the visual representation's encoding of frames showing this activity and further away from the encoding of frames showing other objects like vegetables or unrelated topics like outdoor activities. The semantic closeness can then be measured based on distance metrics (\eg by simple dot product). 

%These relationships can be captured in two ways: first, by training a network that takes pairs of modalities as inputs and predicts a similarity/distance score, or second, by projecting both representations into a joint embedding space such that e.g., the encoding for a sentence like ``Cut the chicken." is close to the encoding of the visual representation of  frames showing this activity and further away from the encoding of  frames showing other objects like vegetables or unrelated topics like outdoor activities \MS{This is very long sentence, break it into a couple}. The semantic closeness can then be measured based on distance metrics. 

Learning such a joint embedding space 
%is usually more efficient than pair-wise distance, especially for large-scale data, but involves the difficult task of learning a respective metric embedding across multiple distinct modalities, as it 
involves the grouping of similar concepts across different modalities. 
Here, it can be helpful to identify which low-level features show activation in certain contexts, which can serve as a form of filtering to focus on relevant inputs and thus learn a good joint embedding space.
Capsule networks \cite{sabour2017dynamic} have been proposed as a technique to capture activations of a specific type of entity and to model higher-level objectness from groups of low-level feature activations. To this end, capsule networks 
%group neural activations/feature outputs and 
find familiar concepts by performing ``high-dimensional coincidence filtering" \cite{hinton2018matrix} through a routing-by-agreement algorithm.
%Furthermore, another novel neural network architecture, capsule networks, have been proposed which actually try to model higher-level objectness from its parts, thus from groups of low-level feature activation. 
%To this end, capsules actually group neural activations/feature outputs and perform a routing-by-agreement algorithm to capture groups of concurrent acivations. \hkc{Is that correct?} 
They have shown their ability in modeling these relationships in images \cite{sabour2017dynamic,hinton2018matrix,kosiorek2019stacked} and videos \cite{duarte2018videocapsulenet}, and have also performed well in multimodal applications \cite{urooj2021found,mcintosh2020visual}. 
However, those approaches have thus far mainly been applied in a fully supervised setting with clean data. 
%Capsules have yet to be applied to noisy unsupervised settings like learning from random YouTube video clips.
%\hkc{we should say here that caps have so far mainly been used in clean fully supervised setting, compared to noise unsupervised data from random youtube clips}

In this work, we leverage the qualities of capsule architectures in the context of multimodal learning to learn a joint embedding space across different input modalities. 
To allow the capsules to learn from large-scale noisy input data, we propose an efficient routing by self-attention mechanism that finds similarity between these lower-level capsule representations to produce higher-level capsules and activations. 
To this end, we build upon the standard capsule network setup and generate a set of capsules for the input features.
From these capsules, we obtain votes for higher-level capsules, in the form of key-query-value tuples, and perform a self attention operation to obtain the higher-level capsule pose representations. These are then passed through a linear layer and a softmax layer to obtain the final activations.
These activations are used to select relevant capsules, increasing the impact of those feature groups belonging to certain object representations while reducing the impact for irrelevant ones.
We find that this self-attention based routing mechanism is more scalable than standard dynamic \cite{sabour2017dynamic} and EM \cite{hinton2018matrix} routing methods: this is vital for applying capsule networks to large-scale video datasets.

%outperforms not only  standard dynamic  approaches, but also standard self-attention applied on low level input features.

%
% Advantages: 
% self attention is more efficient then traditional routing
% it's scalable to capture a large number of concepts 
% performs well
% shared weights work better than separate weights (no experiments, perhaps in supp)
% --- 

%To this end, we build up on the idea a standard capsule setup and use it to learn activation weights for various input features before they are actually projected into the joint embedding space. To allow the capsules to learn such activation weights based on large-scale noisy input data, we propose a new routing by self-attention mechanism that \hkc{...perhaps add some details here ?} 

%It shows that the learned activation actually is able to increase the impact of features belonging to certain object representation and the reduce the impact of no relevant ones. \hkc{not too happy with this sentence ... }

%The multimodal capsule network trained in a self-supervised manner by mapping the capsules with their respective activation to a joint multimodal embedding space which is enforced by the use of a contrastive loss

The proposed multimodal capsule network is trained by mapping the selected capsules to a joint multimodal embedding space which is enforced by the use of a contrastive loss. For evaluation, we train the system on the HowTo100M multimodal dataset, consisting of 1.2 million YouTube instruction videos and evaluate the resulting method on the two zero-shot down-stream tasks of video retrieval on the YouCook2~\cite{zhou2018towards} and MSR-VTT~\cite{xu2016msr} dataset and action localization on the CrossTask~\cite{zhukov2019cross} and the MiningYouTube~\cite{kuehne2019mining} dataset. Our experiments show that the proposed architecture is able to improve performance compared to existing routing mechanisms and to provide competitive performance on all evaluated downstream tasks. 

The contributions of the paper are as follows:
\begin{itemize}
    \item To the best of our knowledge, we are the first to leverage the benefits of capsule architectures for large-scale multimodal data without human annotation.
    \item To this end we propose a novel routing by self-attention mechanism for capsule architectures.
    \item We show that the proposed mechanism is more efficient and scalable than other routing techniques, achieving state-of-the-art results on various challenging benchmark tasks.
\end{itemize}

%\MS{Traditionally before ending introduction section main contributions are summarized.}

% -Our proposed method learns a capsule representation to model the objects in an input \\
% -We then use attention to select the relevant objects \\
% -Then, we map the capsules to a joint multi-modal embedding space which is enforced by the use of a contrastive loss

% -We evaluate our method on several down-stream tasks like zero-shot video retrieval, action localization, and action classification. \\
% -Also, we analyse the learned capsules and attention and show that they learn meaningful object representations

% -------------

% -capsules+routing have learned objectness on visual data well
% - one paragraph talking about different routing mechanisms and how our idea is different (talk about multi-modality)
% --why ours is good: gives good numbers, more efficient than previous routing methods, works on larger data without supervision
% - we want to leverage this for multi-modal self-supervised learning
% - This is how we solve problem
\section{Related Work}

\paragraph{Multimodal Learning}
As annotating large datasets \cite{deng2009imagenet,carreira2017quo}, especially video, is extremely costly, recent approaches started take advantage of the vast amount of video data on websites and social media platforms by leveraging readily available tools like automatic speech recognition systems. This allows the creation of narrated video datasets \cite{miech2019howto100m,sanabria2018how2}. Following that, various methods have been proposed to learn from two or more input modalities, e.g. video-text pairs \cite{amrani2020noise,gabeur2020multi,luo2020univilm,zhu2020actbert,patrick2020support,lei2021less,sun2019videobert,Dong_2019_CVPR}, video-audio pairs \cite{alwassel_2020_xdc,asano2019self,boggust2019grounding,rouditchenko2020avlnet}, or all three modalities, video, audio, and text \cite{alayrac2020self, chen2021multimodal}. 
Most methods use the large-scale data for pretraining the network followed by a fine-tuning on a downstream dataset, which is usually done with less noisy curated or hand-annotated data~\cite{luo2020univilm,patrick2020support,lei2021less,alwassel_2020_xdc,rouditchenko2020avlnet,Dong_2019_CVPR}. However, some approaches show that training on large-scale noisy data alone can also be sufficient and directly apply the model without fine-tuning on the downstream datasets~\cite{amrani2020noise,gabeur2020multi,patrick2020support,sun2019videobert,boggust2019grounding}. 
Especially in the later case, training is usually realized with some variant of a contrastive loss function, such as noise contrastive estimation (NCE) used by \cite{miech2019howto100m} or Masked Margin Softmax (MMS) used by \cite{rouditchenko2020avlnet}. Other methods such as XDC \cite{alwassel_2020_xdc} and MCN \cite{chen2021multimodal} argue that embedding space learning can be improved by also adding a clustering component to the contrastive loss to form groups with similar representations in the embedding space but it usually comes at the cost of having an additional Sinkhorn \cite{cuturi2013sinkhorn} or K-means \cite{lloyd1982least} clustering in the pipeline.
Our proposed method picks up on this idea, but instead of having an explicit clustering objective at the end of the pipeline, we use the implicit characteristics of capsule networks (\ie routing-by-agreement) to group concepts at an earlier stage of the network. 
Technically, most approaches rely on either convolutional architectures \cite{miech2019howto100m, alayrac2020self, chen2021multimodal, amrani2020noise,alwassel_2020_xdc,asano2019self,boggust2019grounding,rouditchenko2020avlnet}, self-attention mechanisms \cite{gabeur2020multi,luo2020univilm,zhu2020actbert,lei2021less,sun2019videobert}  or a combination thereof \cite{patrick2020support}.
%, with self-attention based methods mostly relying on much larger pretraining corpora than their convolutional counter parts. 
%
To the best of our knowledge, this work makes a first attempt, to employ the abilities of capsule networks to address the problem of multimodal self-supervised learning. The proposed approach uses self-attention as an efficient routing mechanism to attend to relevant elements (\ie capsules) in large-scale data. 
\paragraph{Capsule Networks}
The concept of capsule networks was first introduced in \cite{hinton2011transforming}, where view-equivariant vector representations were learned from images. Sabour \etal \cite{sabour2017dynamic} extended this idea and proposed an iterative routing-by-agreement algorithm which was able to classify and segment overlapping digits. Capsule networks have been widely applied to various domains and problems including text classification \cite{yang2018investigating}, video action detection \cite{duarte2018videocapsulenet}, point cloud processing \cite{zhao20193d}, and medical image segmentation \cite{lalonde2021capsules}. One key aspect of capsules networks is their ability to route and hierarchically activate higher-level capsules based on agreement of multiple lower-level capsules. However, this ability comes with the increased computational cost of the routing-by-agreement algorithm.
First capsule architectures~\cite{sabour2017dynamic} used dynamic routing which can be computationally slow and results in high memory consumption, especially on higher dimensional input data and for a larger number of capsules.
Hinton \etal \cite{hinton2018matrix} reduced the number of parameters by learning matrix capsules with an iterative Expectation-Maximization (EM) based routing algorithm. 
Several other works have attempted to make more efficient and scalable routing algorithms including self-routing capsule networks \cite{hahn2019self}, KDE-based routing \cite{zhang2018fast}, STAR-Caps \cite{ahmed2019star}, spectral capsule networks \cite{bahadori2018spectral}, and subspace capsule networks \cite{edraki2020subspace}. 
% Recently, Kosiorek \textit{et al.} \cite{kosiorek2019stacked} proposed using Set Transformer \cite{lee2019set} as an efficient routing procedure to discover objects from their constituent parts.
%
%One key aspect of capsules is their ability to route, thus to activate higher level capsules based on the agreement of multiple lower level capsules.
%
% Therefore, e.g. dynamic routing involves a squashing procedure to map features to a lower dimensional space to allow for a computationally feasible training. 
% Practically, we observe that the EM routing can become unstable when trained on noisy data. Additionally, these types of capsules have, so far, mainly been used in the context of fully supervised settings. 
%
%
Recently, Efficient-CapsNet \cite{mazzia2021efficient} and stacked capsule autoencoders (SCAEs) \cite{kosiorek2019stacked} have proposed routing mechanisms based on attention. Efficient-CapsNet uses vector capsules similar to those found in \cite{sabour2017dynamic}, and computes self-attention across votes of the lower-level capsule layers to find the routing/coupling coefficients. Then, the resulting higher-level capsules are a weighted sum across these same votes based on the coefficients.  Contrarily, our proposed self-attention routing method generates separate key, query, and value representations. This allows us to separate the computation of the votes and the routing coefficients: the key and query generate the routing coefficients, which are used to weight the votes to obtain the higher-level capsules. SCAEs adapt the Set Transformer \cite{lee2019set} to perform routing between the set of part capsules to the object capsules. They generate capsule activations by maximising the part pose likelihood from a mixture of predictions from lower-level capsules. 
%\MS{say here why previous methods on capsule attention are not applicable to multi-modal?}. 
%Instead of utilizing a Set Transformer, we propose to use self-attention as a routing mechanism.% and generate the query by linear projection from the input capsules.
%to adapt the concept of attention-based routing to the problem of multimodal learning from large-scale noisy data, . 
Different from SCAE, we propose to use self-attention as a routing mechanism and we compute activation probabilities by the linear transformation of the higher level features followed by a softmax. We find this setting to be computationally more efficient allowing us to train without the need for tuning of sparsity constraints, while showing higher performance compared to Set Transformer in the targeted setup (see Sec. \ref{sec:experiments_ablation}). 

%Recently, a mechanism also based on attention, Set Transformer, has been proposed by \cite{lee2019set} and adapted for routing in context on capsules for stacked capsule autoencoders by \cite{kosiorek2019stacked}.
%While Set Transformer mainly focuses on an attention-based neural network that processes sets of data, stacked capsule autoencoders explicitly use this technique in the context of capsules and show that such routing can be useful on datasets like MNIST and Cifar-10. 
%Furthermore, stacked capsule autoencoders generate activations by maximising the part pose likelihood from a mixture of predictions from lower-level capsules. 
%To adapt the concept of attention-based routing to the problem of self-supervised multimodal learning from large-scale noisy data, we propose to use self-attention instead of Set Transformer and generate the query by linear projection from the input capsules. Additionally, different from stacked capsule autoencoders, we solely rely on the self-attention mechanism for routing and compute activation by the linear transformation of the higher level features followed by a softmax. We found this setting is computationally more efficient and allows us to train without the need for tuning of sparsity constraints, while showing higher performance compared in the targeted setup (see Sec. \ref{sec:experiments_ablation}).

\section{Multimodal Learning with Capsule Networks}
\label{sec:method}
\begin{figure*}[t]
    \includegraphics[width=\textwidth]{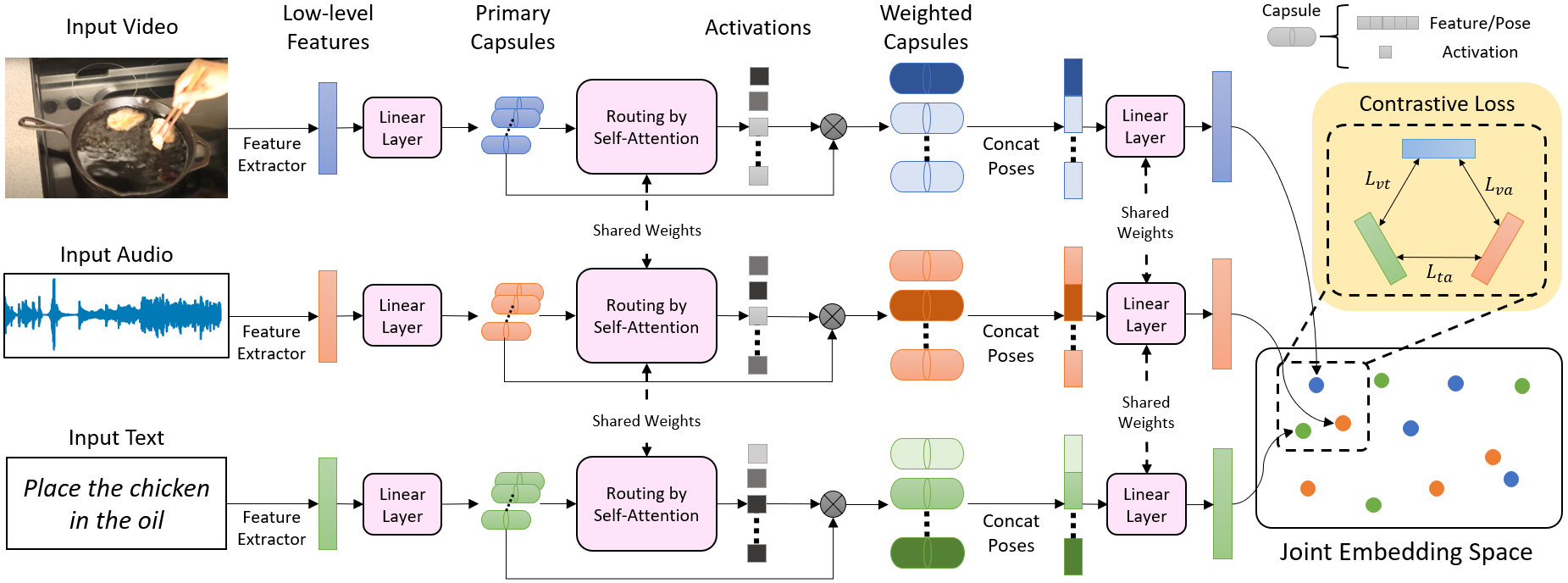}
    \caption{\textbf{Overview of our proposed approach.} Given a video, audio, and text triplet, the network extracts modality specific features and converts them into a set of primary capsules. Then, these capsules are  routed using self-attention to obtain a higher-level activations, which are used to weight capsule features. The weighted capsule features are projected into a final joint multimodal feature representation. This joint representation space is enforced by a pair-wise contrastive loss. }
    \label{fig:architecture}
    \vspace{-2mm}
\end{figure*}
%\MS{This much nicer figure compared to previous version. Instead of weighted capsules do you want to say secondary capsules? Also, it will be good to put the capsules in a block to highlight them, because those are the main contribution}

In the following, we first describe the proposed  multimodal capsule network architecture (Figure \ref{fig:architecture}) at high level, and then follow with a detailed description of the proposed routing by self-attention mechanism (Figure \ref{fig:selfatt}). We close with a description of the training procedure.  
%We close with a short discussion on how the proposed routing mechanism complements the set of existing routing approaches.
\vspace{-4mm}
\paragraph{System setup}
%Given $n$ video clips, each with a corresponding video representation, audio representation, and text narration we attempt to learn a joint multimodal representation space. 
Given $n$ video clips, each with a corresponding video, audio, and text representations we attempt to learn a joint multimodal representation space. We denote the video as $v\in\mathcal{V}$, the audio as $a\in\mathcal{A}$, and the text narration generated by an automated speech recognition (ASR) system as $t\in\mathcal{T}$. Thus, the training set of $n$ video clips is representated by tuples $\{\left(v_i, a_i, t_i\right)\}_{i=1}^{n}$. Contrastive multimodal learning attempts to learn a set of functions to generate embeddings for each modality such that embeddings for semantically similar inputs are closer together than semantically dissimilar inputs. Formally, we learn functions, $f_v: \mathcal{V} \rightarrow \mathbb{R}^D$, $f_a: \mathcal{A} \rightarrow \mathbb{R}^D$, and $f_t: \mathcal{T} \rightarrow \mathbb{R}^D$ which create $D$ dimensional embeddings (i.e. $f_v\left( v\right) \in \mathbb{R}^D$). The input representations take the form of pre-extracted 2D and 3D features from a video clip, log-mel spectrograms extracted from an audio segment, and a text embedding extracted by sentence-based neural network. The goal is to find mapping functions $f_v$, $f_a$, and $f_t$, so that the distance of all possible pairs from the same tuple $(v_i,a_i)$, $(t_i,a_i)$, and $(v_i,t_i)$ is minimized in the embedding space and the distance to all other tuple pairs is maximized. An overview of the overall system is shown in Figure \ref{fig:architecture}.

\vspace{-1mm}
\subsection{Multimodal Capsule Architecture}

\begin{figure*}[t]
    \includegraphics[width=\textwidth]{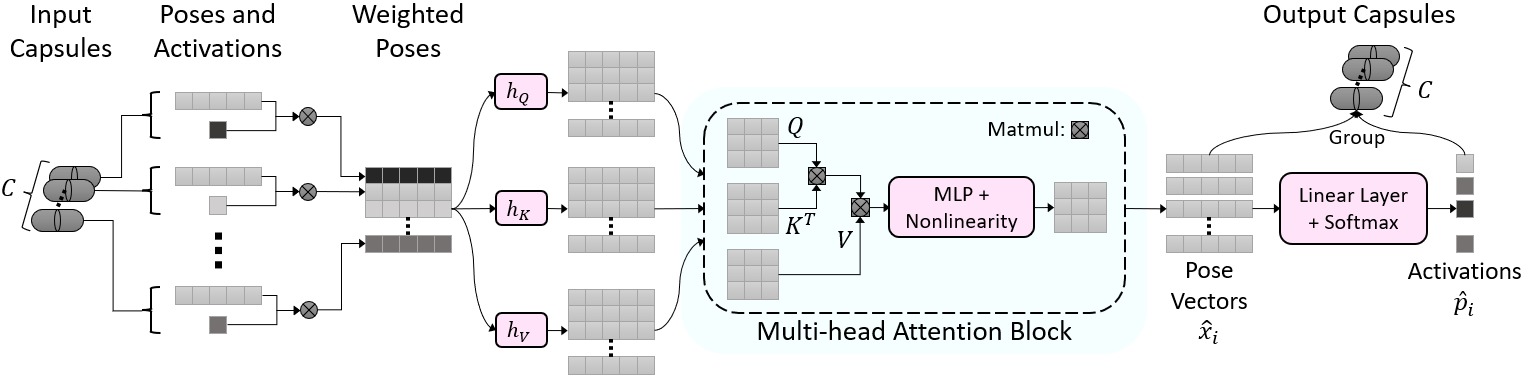}
    \caption{\textbf{Proposed Routing by Self-Attention.} The input is a set of $C$ capsules. The activation-weighted capsule features are projected into query, key, and value matrices which are used in a multi-head self-attention block to generate higher-level capsule poses. A linear transformation with softmax activation then generates the activations for these higher-level capsules.}
    \label{fig:selfatt}
    \vspace{-2mm}
\end{figure*}

\paragraph{Primary Capsules} To learn the mapping of each input feature to the joint embedding space, i.e. functions $f_v$, $f_a$, and $f_t$, we propose a novel capsule network architecture. 
%A capsule is a group of neurons representing an entity or part of an entity.
%Relationships between different entities and their parts are modeled through a routing-by-agreement algorithm. 
A capsule is composed of a multi-dimensional pose vector $x$, which represents an entity's properties and an activation $p$, which represents an existence probability (i.e. the probability that the given entity/object exists within the input).
From each input modality feature, a learned linear layer extracts a set of $C$ primary capsule poses; the $C$ activations are produced by a linear layer followed by a sigmoid non-linearity. 
% As the routing algorithm, we use self-attention \cite{vaswani2017attention} to learn relationships between capsules.
We denote the $i$-th capsule for modality $m$ has the pose vector $x_i^m\in \mathbb{R}^{d_1}$ and activation $p_i^m\in[0,1]$. 
We use these capsules in a self-attention based routing-by-agreement algorithm, depicted in Figure \ref{fig:selfatt}, to learn the relationships between the entities they model. 
%--
\vspace{-2mm}
\paragraph{Routing by Self-Attention}
Routing is the method by which capsule networks propagate information from one layer to the next. It is a type of voting mechanism which obtains votes from capsules in a lower capsule layer, and finds agreement between these votes which are aggregated into capsules in the higher capsules layer. To this end, traditional routing approaches (\eg dynamic and EM routing) find agreement in high-dimensional space so that sets of lower level capsules can vote for higher-level capsules. This process is computationally expensive as it often relies on an iterative procedure, and has therefore mainly been used on small-scale data. We propose to replace this routing mechanism by utilizing self-attention, as self-attention can provide a similar mechanism for finding agreement between high-dimensional vectors.

Practically, we first multiply capsule pose vectors $x_i^m$ by their respective activations $p_i$ to ensure entities which are not present (i.e. $p_i^m \rightarrow 0$) are not used in the routing process. We then learn a set of functions to extract the respective key, query, and value representations from capsules $K=h_K( p_i^m x_i^m ),Q=h_Q( p_i^m x_i^m ),V=h_V( p_i^m x_i^m )$, using two linear layers for all functions $h$. 
% We propose  a novel  self-attention based routing algorithm which takes into account the existing probabilities of the capsules. 
These learned functions, $h_K,h_Q,h_V:\mathbb{R}^{d_1}\rightarrow \mathbb{R}^{d_2}$, map the primary capsules pose vectors to the secondary capsules' pose feature space and are used in a multi-head self-attention mechanism:
\begin{equation}
\small
    \hat{x}_i^m = \text{Attention}\left(Q, K, V\right) = g\left(\text{softmax}\left(\frac{QK^T}{\sqrt{d_2}}\right)V\right),
\end{equation}
where $g$ is a two-layer nonlinear MLP\footnote{See Appendix for additional details}. In the context of capsule routing, the query and key produce the routing coefficients which determine the amount of information a lower-level capsule sends a specific higher level capsule, whereas the value can be considered a vote, or prediction, for the properties of the higher level capsule. 
% ---
%  and are Here, the pose vectors are multiplied by the activations to ensure objects which are not present (i.e. $p_i^m \rightarrow 0$) are not used in the routing process.

From the secondary capsule layer's poses, $\hat{x}_i^m$, we obtain their existence probabilities, through a softmax operation:
\begin{equation}
\label{eq:acts}
\small
    \hat{p}_i^m =  \frac{\exp\left(x_i^m W_p + b_p\right)}{\sum_{j=1}^{C}\exp\left(x_i^m W_p + b_p\right)},
\end{equation}
where $W_p\in\mathbb{R}^{d_2\times1}$ and $b_p\in\mathbb{R}$ are learned parameters. These probabilities are used to select relevant capsules. 

By analysing these existence probabilities for different inputs, we find that these higher-level capsules seem to capture and weight the existence of certain concepts at different granularity levels: some capsules represent general concepts  ``outdoors" or ``video games", while others are more specific like ``paper folding" or ``putting things in bowls". A detailed discussion can be found in Section \ref{sec:experiments_qualitative}.

%Looking at the existence probabilities in secondary capsules for different input, we found that in the here addressed case of multi-modal learning secondary capsules seem to capture and weight the existence of certain concepts at different granularity levels such as outdoor or video games but also paper folding or putting thing in bowls etc. A detailed discussion can be found in Section \ref{sec:experiments_qualitative}.

% ---
\paragraph{Mapping to Joint Embedding Space} To finally map all capsules to a joint embedding space, we concatenate the activation-weighted capsules to a single vector and pass them through a linear transformation, $f_\text{out}$ to obtain the final feature representations:
\begin{equation}
\small
    f_v =  f_\text{out}\left(\hat{p}_i^v x_i^v\right), f_a =  f_\text{out}\left(\hat{p}_i^a x_i^a\right), \text{ and } f_t =  f_\text{out}\left(\hat{p}_i^t x_i^t\right).
\end{equation}
Note that all learned weights used after the generation of the primary capsule layer, namely $h_K,h_Q,h_V$ and $f_\text{out}$, are shared across modalities. 
% We find this aids in the learning of a final joint representation space. \hkc{Last sentence would only make sense if we could somehow show this in the evaluation section}

\subsection{Contrastive Multimodal Learning}
To train the described architecture and learn the joint representation space, we use a contrastive loss on each pair of modalities $(v,a)$, $(t,a)$, and $(v,t)$. For different modalities from the same video clip, the contrastive loss maximizes the similarity of their embeddings; conversely, it  minimizes the similarity for embeddings from different video clips. Following \cite{rouditchenko2020avlnet}, we use the Masked Margin Softmax (MMS) loss \cite{ilharco2019large}, which defines the dot-product between two vectors as the similarity measure and computes similarities across a batch of $B$ samples. 

The loss is computed between two modalities, and can be viewed as the sum of two instances of InfoNCE~\cite{oord2018representation} (with a margin $\delta$). For example, the loss for the visual/audio pair ($L_{va}$) consists of two components: the first where the visual input is fixed and audio samples are varied, and the second where the audio input is fixed and visual samples are varied.
For the video-audio pair the loss is defined as:
\iffalse
\begin{equation}
\tiny
    L_{va} = \frac{1}{B} \sum_{i=1}^B\left[ \log \frac{e^{f_v\left(v_i\right)\cdot f_a\left(a_i\right)-\delta}}{e^{f_v\left(v_i\right)\cdot f_a\left(a_i\right)-\delta} + \sum\limits_{\substack{j=1 \\ j\neq i}}^{B}e^{f_v\left(v_j^\text{imp}\right)\cdot f_a\left(a_i\right)}} +  \log \frac{e^{f_v\left(v_i\right)\cdot f_a\left(a_i\right)-\delta}}{e^{f_v\left(v_i\right)\cdot f_a\left(a_i\right)-\delta} + \sum\limits_{\substack{k=1 \\ k\neq i}}^{B}e^{f_v\left(v_i\right)\cdot f_a\left(a_k^\text{imp}\right)}}\right],
\end{equation}
\fi
%\tiny
\begin{multline}
    L_{va} = \frac{1}{B} \sum_{i=1}^B\log \frac{e^{f_v\left(v_i\right)\cdot f_a\left(a_i\right)-\delta}}{e^{f_v\left(v_i\right)\cdot f_a\left(a_i\right)-\delta} + \sum\limits_{\substack{j=1 \\ j\neq i}}^{B}e^{f_v\left(v_j^\text{imp}\right)\cdot f_a\left(a_i\right)}} + \\
    \log \frac{e^{f_v\left(v_i\right)\cdot f_a\left(a_i\right)-\delta}}{e^{f_v\left(v_i\right)\cdot f_a\left(a_i\right)-\delta} + \sum\limits_{\substack{k=1 \\ k\neq i}}^{B}e^{f_v\left(v_i\right)\cdot f_a\left(a_k^\text{imp}\right)}},
\end{multline}
%\normalsize
%\MS{why this equation appears ugly?}
where $\delta$ is an empirically selected hyper-parameter. Here, $v_j^\text{imp}$ and $a_k^\text{imp}$ are ``imposters" (i.e. samples from the batch and do not co-occur within the same time-frame) from the video and audio modalities, respectively. For this video-audio case, the loss discriminates between the true embedding pairs $\left(v_i,a_i\right)$ and the imposter pairs $\left(v_k^\text{imp},a_i\right)$ and $\left(v_i,a_j^\text{imp}\right)$, for all $k\neq i$ and $j\neq i$ in the batch. 
We sample negatives from both within the same video and from other videos, since this has been shown to empirically improve performance~\cite{miech2019howto100m}.
%\MS{do you want to say why did you pick this loss instead of standard contrastive loss and will you provide comparison in the supplementary material?}
The final loss is the sum of the pairwise MMS losses between different modalities:
\begin{equation}
    L_\text{final} = L_{va} + L_{ta} + L_{vt}.
\end{equation}
Since the loss is computed over all modality pairs, it ensures all features are projected into the same space and are comparable.

\section{Experimental Evaluation}
\label{sec:experiments}
In this section, we assess the performance of the proposed approach in the context of multimodal learning. For this evaluation, we focus on the zero shot capabilities of the proposed approach, namely on the downstream tasks of zero-shot text-to-video retrieval and zero-shot temporal action localization, as this allows us to evaluate how well high-level semantic concepts  have been identified and grouped across various modalities. We first present an overview on the implementation details of our proposed approach. The overall system performance is then compared with various other techniques in the field. We finally evaluate the impact of each component including the routing mechanism in comparison with other available techniques and present qualitative results for the proposed method. The code and related resources is publicly available at \url{https://github.com/KevinDuarte/Multimodal-Capsule-Networks}.

%We start with the implementation detail, then first compare the overall system performance to other approaches in the field and second, evaluate in impact of its components including routing compared to other available techniques.    

\subsection{Implementation Details} 
%To demonstrate our method's ability to generalize well across different visual features, we evaluate with two different visual backbones. First, f
Following \cite{miech2019howto100m}, the input visual features for our method are 2D features extracted at 1 fps using a ResNet-152 model \cite{he2016deep} pretrained on ImageNet \cite{deng2009imagenet}, as well as 3D features extracted at 1.5 fps using a ResNext-101 \cite{hara2018can} pretrained on Kinetics \cite{carreira2017quo}. These features are max-pooled over time and concatenated to form a 4096 dimension feature vector for a given video clip. %We also evaluate using the S3D-G \cite{xie2018rethinking} backbone pretrained by \cite{miech2020end}, following the UniVL \cite{luo2020univilm} setup.
%Following \cite{miech2019howto100m}, the input visual features for our method are 2D features extracted at 1 fps using a ResNet-152 model \cite{he2016deep} pretrained on ImageNet \cite{deng2009imagenet}, as well as 3D features extracted at 1.5 fps using a ResNext-101 \cite{hara2018can} pretrained on Kinetics \cite{carreira2017quo}. These features are max-pooled over time and concatenated to form a 4096 dimension feature vector for a given video clip.
The audio input to our network are features extracted from the log-mel spectrograms by a pre-trained DAVEnet model \cite{harwath2018jointly}. For textual features, we follow \cite{miech2019howto100m}, and use a GoogleNews pretrained Word2vec model \cite{mikolov2013efficient} to extract word embeddings. Then a max-pooling operation over all word embeddings in a sentence produces a single vector representation. All feature extraction backbones are fixed (i.e. not fine-tuned) during training and evaluation. To train our network we use an Adam optimizer \cite{kingma2014adam} with a learning rate of 0.001 and cosine learning rate scheduler \cite{misra2020self}. The model is trained on 4 V100 GPUs for 20 epochs, using a batch size of 4096. In the MMS loss, we set $\delta=0.001$. Unless otherwise stated, we set the number of capsules to $C=128$, the dimension of each capsule's pose vector to $d_1=32$ and $d_2=256$, and the final joint representation dimension is $D=4096$. Our method is trained using the HowTo100M \cite{miech2019howto100m} instructional video dataset, which consists of 1.2 million videos with corresponding audio and text transcripts extracted using an off-the-shelf ASR system. The video-audio-text tuples are defined by the transcription timestamps provided with the dataset. 

\subsection{Text-To-Video Retrieval}

\begin{table}
    \tablestyle{3pt}{1.05}
    \caption{Evaluation of zero-shot text-to-video retrieval. MIL-NCE* uses the same training procedure as \cite{miech2020end} with different backbone features, $\dagger$ indicates trainable backbone. Modality indicates the modalities used during inference, where V: video, T: text, A: audio.
    \label{tab:retrieval}
    }
    \resizebox{1\columnwidth}{!}{
    \begin{tabular}{@{}l|c|c|cccc|cccc@{}}
    	\toprule
    	\multicolumn{3}{c}{} & \multicolumn{4}{c}{YouCook2} & \multicolumn{4}{c}{MSR-VTT} \\ 
    	\cmidrule(lr){4-7} \cmidrule(lr){8-11} 
    	  Method & Modality & Visual Backbone & R@1$\uparrow$  & R@5$\uparrow$ & R@10$\uparrow$ & MedR$\downarrow$ & R@1$\uparrow$ & R@5$\uparrow$  & R@10$\uparrow$ & MedR$\downarrow$\\ 
    	\midrule
        MMT~\cite{gabeur2020multi} & VT & 7 experts & - & - & - & - & - & 14.4 & - & 66 \\
        %NoiseEstimation \cite{amrani2020noise} & VT & R152 & - & - &- & -& 8.4 & 22.0 & 30.4 & 36 \\
    	ActBERT \cite{zhu2020actbert} & VT & R101+Res3D & 9.6 & 26.7 & 38.0 & 19 & 8.6 & 23.4 & 33.1 & 36 \\
    	Support Set \cite{patrick2020support} & VT & R152+R(2+1)D-34 & - & - & - & - & 8.7 & 23.0 & 31.1 & 31 \\ 
    	MIL-NCE~\cite{miech2020end}$\dagger$ & VT & I3D-G & 11.4 & 30.6 & 42.0 & 16 & 9.4 & 22.0 & 30.0 & 35 \\
        MMV FAC \cite{alayrac2020self}$\dagger$ & VAT & TSM-50x2 & 11.7 & 33.4 & 45.4 & 13 & 9.3 & 23.0 & 31.1 & 38 \\
        CLIP4Clip \cite{amrani2020noise} & VT & CLIP & - & - &- & -& 31.2 & 53.7& 64.2 & 4 \\
        \midrule
    	HT100M \cite{miech2019howto100m} & VT &  R152+RX101 & 6.1 & 17.3 & 24.8 & 46  & 7.2 & 19.2 & 28.0 & 38     \\
    	NoiseEstimation \cite{amrani2020noise} & VT & R152+RX101 & - & - &- & - & 8.0 & 21.3 & 29.3 & 33 \\
    	MIL-NCE* \cite{miech2020end} & VT & R152+RX101 & 8.0 & 22.9 & 32.1 & \textbf{29} & 8.6 & 23.1 & \textbf{30.8} & 33 \\
    	% AVLNet \cite{rouditchenko2020avlnet} & VT &  R152+RX101 & - & - & - & - & - & - & - & - \\
    	Ours & VT & R152+RX101 & \textbf{9.0} & \textbf{23.2} & \textbf{32.5} & 30 & \textbf{9.7} & \textbf{23.2} & 30.7 & \textbf{32} \\
        \midrule
        FC Baseline & VAT &  R152+RX101 & 15.5 & 31.0 & 40.0 & 22 & 7.8 & 17.8 & 25.2 & 50 \\
    	Self-Attention Baseline & VAT &  R152+RX101 & 13.8 & 29.3 & 36.2 & 34 & 8.9 & 22.3 & 30.1 & 41 \\
    	AVLNet \cite{rouditchenko2020avlnet} & VAT &  R152+RX101 &  \textbf{19.9} & 36.1 & 44.3 & 16 & 8.3 & 19.2 & 27.4 & 47 \\
    	MCN \cite{chen2021multimodal} & VAT & R152+RX101 & 18.1 & 35.5 &  45.2 & - & \textbf{10.5} & \textbf{25.2} &  \textbf{33.8} & - \\
    	Ours & VAT & R152+RX101 & 19.3 & \textbf{37.8} & \textbf{47.3} & \textbf{13} & 9.6 & 22.6 & 32.0 & \textbf{32}  \\
    	%R@1: 0.0870 - R@5: 0.1930 - R@10: 0.2830 - Median R: 40.0
         \midrule
    	MIL-NCE~\cite{miech2020end}$\dagger$ & VT & S3D-G & 15.1 & 38.0 & 51.2 & 10  & 9.9 & 24.0 & 32.4 & 29.5 \\
    	Ours & VT & S3D-G & 14.7 & 36.2 & 47.9 & 12 & 8.3 & 19.0 & 27.5 & 42 \\
    	Ours & VAT & S3D-G & 23.0 & 43.3 & 53.8 & 8 & 8.7 & 19.3 & 28.3 & 40 \\
    	\bottomrule
    \end{tabular}
    }
    \vspace{-0.2cm}
\end{table}

\paragraph{Datasets and Metrics} The problem of text-to-video retrieval involves searching a pool of videos for a single video that corresponds to a given ground-truth text query.  We evaluate zero-shot text-to-video retrieval on the YouCook2 \cite{zhou2018towards} and MSR-VTT \cite{xu2016msr} datasets, which are common benchmark datasets for zero-shot video retrieval. The YouCook2 dataset consists of cooking instructional video clips with human-annotated text descriptions, and we use the validation set of 3.5k clips following prior work \cite{miech2019howto100m,miech2020end}. The MSR-VTT dataset contains 10K video clips with human-annotated captions on various topics, and we use the test set of 1K video clips from \cite{miech2019howto100m}. 
%Note that the human generated captions in both datasets deviate for the original training domain of ASR text, which is usually more noisy and differs from the structure of written out, descriptive sentences. 
For the retrieval task, we  compute the euclidean distance between the text and video representations through the pretrained network to find e.g. the top video candidates for a given text sample. For both datasets, we report the recall metrics: R@1, R@5, R@10, and Median Recall (MedR).

\vspace{-2mm}
\paragraph{Comparison with the state-of-the-art} We report the results on the text-to-video retrieval task for YouCook2 and MSR-VTT in Table \ref{tab:retrieval} for two cases, zero-shot text-to-video retrieval (VT) and zero-shot text-to-video+audio retrieval (VAT). 
%For a fair comparison, the first part presents approaches that use the same backbone and do not fine-tune this backbone during training, the second part of the table summarizes all other approaches which either use a different backbone or fine-tune the backbone during training or both.
%with the S3D-G backbone, our capsule-based approach obtains improvements of 6.8\% on R@1 and 3.6\% on R@5. 
%Furthermore, it achieves comparable results to models with trainable backbones, with an improvement of 4.2\% on R@1. 
We find that our method achieves strong performance when compared with prior approaches
%which use the same video backbone 
in both cases and on both downstream datasets. %, while the improvement on MSR-VTT is not as pronounced as the improvements achieved on YouCook2.
%On MSR-VTT, however, we find that our approach achieves slightly lower performance than previous methods.
%Comparing results on both datasets, it can be assumed that the model has actually captured the various concepts present in instructional videos as they also arise in YouCook2, but struggles with the domain shift between the instructional video domain of HowTo100M and the audio and text descriptions in MSR-VTT, which are not instructional. %\MS{How do I infer all of this just looking at the Table 1?}.
Notably, the addition of the audio modality leads to a large performance boost on YouCook2. However, we observe that systems trained in a multimodal setup may be more sensitive to domain shifts as compared to single modality tasks like image classification. As we are dealing with multiple modalities, the misalignment of only a single modality from the training data  may lead to reduced performance of the model. This can be seen in the comparison of the performance of different modality combinations of MSRVTT. Yet, in this case, better performance can be reach by simply dropping the misaligned domain. On the other hand, when all modalities more closely resemble the training data, as is the case with YouCook2, there is generally a performance improvement.

\begin{table}
% 		\tablestyle{2pt}{1.05}
        \caption{Evaluation of zero-shot temporal action localization. MIL-NCE* uses the same training procedure as \cite{miech2020end} with different backbone features, $\dagger$ indicates trainable backbone. %Modality indicates the modalities used during inference, where V: video, T: text, A: audio.
		\label{tab:temporal_state}
		%\vspace{-0.2cm}
		}
		\resizebox{\columnwidth}{!}{
		\begin{tabular}{@{}l|c|ccc|ccc@{}}
            \toprule
            \multicolumn{2}{c}{}& \multicolumn{3}{c}{CrossTask} & \multicolumn{3}{c}{MYT}   \\ 
            \cmidrule(lr){3-5} \cmidrule(lr){6-8} 
            Method & Visual Backbone & Recall$\uparrow$ & IOD$\uparrow$ & IOU$\uparrow$ & Recall$\uparrow$ & IOD$\uparrow$ & IOU$\uparrow$ \\ 
            \midrule
            Cross-task (superv.) \cite{zhukov2019cross} & R152+I3D & 31.6  & - & - & -  & - & -  \\
            Cross-task (weakly superv.) \cite{zhukov2019cross} & R152+I3D & 22.4 & - & -  & -  & - & -  \\
            ActBERT~\cite{zhu2020actbert} & R101+Res3D & 37.1 & - & -   &  - & - & - 
            \\
            ActBERT~\cite{zhu2020actbert} & + Faster R-CNN& 41.4 & - & -   &  - & - & - \\
            MIL-NCE~\cite{miech2020end}$\dagger$ & I3D-G & 36.4 & - & -   & - & - & -                      \\
            
            Mining: GRU (superv.) \cite{kuehne2019mining} & TSN & -  & - & -  &   - & 14.5 & 7.8  \\
            Mining: MLP (weakly superv.) \cite{kuehne2019mining} & TSN & -  & - & -   & - & 19.2 & 9.8  \\
            \midrule
            HT100M~\cite{miech2019howto100m} & R152+RX101 & 33.6 & 26.6 & 17.5   & 15.0 & 17.2 & 11.4 \\
			MIL-NCE*~\cite{miech2020end} & R152+RX101 & 33.2 & 30.2 & 16.3   & 14.9 & 26.4 & 17.8  \\
			MCN \cite{chen2021multimodal} & R152+RX101 & 35.1 & 33.6 & 22.2 & 18.1 & 32.0 & 23.1 \\
            Ours & R152+RX101 & 35.2 & 32.6 & 21.4 & 18.0 & 31.6 & 22.9 \\
            
            \midrule
            MIL-NCE~\cite{miech2020end}$\dagger$ & S3D-G & 40.5 & - & -   & - & - & -                      \\
			UniVL \cite{luo2020univilm}$\dagger$ & S3D-G & 42.0 & - &  -  & - & - & - \\
			Ours & S3D-G & 43.2 & 35.4 & 23.1 & 22.1 & 34.2 & 25.4 \\
			\bottomrule
		\end{tabular}
		}
        \vspace{-0.4cm}
\end{table}

\subsection{Temporal Action Localization}
%\MS{It will be good to explain how exactly you do action localization once you learn representation in self-supervised manner; the standard reader will be lost here. In this piece you do not use any text and audio? Even though the representation is learned through these modalities as well?}
\paragraph{Datasets and Metrics} Given a set of action classes, the goal of temporal action localization is to predict the actions present at each time-step of the video. 
In this task, we compute the distance between the video representation and each action's text representation to obtain a class prediction for each time-step of the video.
We evaluate on the CrossTask \cite{zhukov2019cross} and Mining YouTube \cite{kuehne2019mining} datasets. CrossTask contains 2.7k instructional videos; each video frame is manually annotated using action steps/ordering for each task collected from \textit{wikiHow}. The recall is calculated using the same inference procedure of \cite{zhukov2019cross}. The Mining YouTube dataset contains videos from five simple cooking recipes - ``eggroll", ``fried egg", ``pancake", ``omelet", and ``scrambled egg". The test set contains 50 videos from each task (250 in total) that are densely annotated with 512 classes comprised of verb-object pairs (94 unique verbs and 171 objects). For evaluation, we report the recall metric as well as the intersection over detection (IoD) \cite{bojanowski2014weakly} and intersection over union (IoU) metrics as outlined in \cite{kuehne2019mining}. 
%The IoD and IoU metrics are defined as $\frac{G\cap D}{D}$ and $\frac{G\cap D}{G\cup D}$, respectively, where $G$ is the ground-truth action and $D$ is the prediction.
The IoD metric is defined as $\frac{G\cap D}{D}$ and the IoU metric is defined as $\frac{G\cap D}{G\cup D}$, where $G$ is the ground-truth action and $D$ is the prediction.

\vspace{-4mm}
\paragraph{Comparison with the state-of-the-art} We present the results for the temporal action localization task in Table \ref{tab:temporal_state}. When compared to methods with the R152+RX101 backbone feature extractor \cite{miech2019howto100m,miech2020end,chen2021multimodal}, we show strong performance across both datasets and all metrics. On CrossTask, our proposed method achieves improved recall compared to both MIL-NCE~\cite{miech2019howto100m} and UniVL~\cite{luo2020univilm} when using the S3D-G backbone.
%MIL-NCE~\cite{miech2019howto100m} achieves improved recall with stronger backbone features and ActBERT~\cite{zhu2020actbert} uses a stronger language model as well as region-based features extracted by a Faster R-CNN. 
Furthermore, our method outperforms the fully supervised baseline in \cite{zhukov2019cross} and the state-of-the-art weakly supervised approach \cite{kuehne2019mining} on the reported metrics in CrossTask and Mining YouTube, respectively.

%\vspace{-2mm}
\subsection{Ablations}

Here, we present ablations to evaluate our proposed self-attention based routing mechanism's efficacy, its ability to scale with more capsules, and compare with other architectural baselines. Additional ablations are contained in the appendix. %Appendix \ref{sec:app:ablations} contains additional ablations.

\begin{table}
    % \tablestyle{2pt}{1.05}
    \caption{Evaluation of different types of routing functions as well as without routing for $C=64$ number of capsules and a dimensionality of $d_1=d_2=16$ including runtime and memory usage.
    \label{tab:ablation-routing}
    \vspace{-0.5cm}
    }
    \centering
    \resizebox{1\columnwidth}{!}{
    \begin{tabular}{@{}l|cc|cc|c|c@{}}
    	\toprule
    	\multicolumn{1}{c}{} & \multicolumn{2}{c}{YouCook2} & \multicolumn{2}{c}{MSRVTT}  \\ 
    	\cmidrule(lr){2-3} \cmidrule(lr){4-5} 
    	Method  & R@1  & R@10 & R@1  & R@10 & Memory Usage (GB) & Run-time (sec/batch) \\ 
    	\midrule
    	No Routing & 15.3 & 41.9 & 7.6 & 30.1 & 9.12 & 0.687 \\
    	\midrule
    	Dynamic Routing \cite{sabour2017dynamic} & 17.0 & \textbf{44.3} & 8.2 & 31.1 & 20.50 & 1.534 \\
    	EM Routing \cite{hinton2018matrix} & 5.8 & 24.2 & 5.7 & 21.8 & 19.13 & 1.272  \\
    	Set Transformer \cite{lee2019set} & 16.5 & 40.0 & 8.4 & 30.0 & 9.11 & 0.707  \\
    	Self-Attention (ours) & \textbf{18.6} & 44.0 & \textbf{8.7} & \textbf{31.6} & 9.11 & 0.722\\
    	\bottomrule
    \end{tabular}
    }
    %\vspace{-0.25cm}
\end{table}

\begin{table}
    % \tablestyle{2pt}{1.05}
    \caption{Evaluation on different number of capsules for a dimensionality of $d_1=32$ and $d_2=256$. It shows that on the given dataset we reach saturation around $C=128$ capsules.
    \label{tab:ablation-numcaps}
    \vspace{-0.2cm}
    }
    \centering
    \resizebox{1\columnwidth}{!}{
    \begin{tabular}{@{}l|cc|cc|c|c@{}}
    	\toprule
    	\multicolumn{1}{c}{} & \multicolumn{2}{c}{YouCook2} & \multicolumn{2}{c}{MSRVTT}  \\ 
    	\cmidrule(lr){2-3} \cmidrule(lr){4-5} 
    	Method  & R@1  & R@10 & R@1  & R@10 & Memory Usage (GB) & Run-time (sec/batch) \\ 
    	\midrule
    	$C=32$  & 18.5 & 45.0 & 8.0 & 29.2 & 9.15 &  0.730 \\
    	$C=64$  & 18.1 & 46.1 & 8.6 & 29.4 & 11.24 & 0.768 \\
    	$C=128$ & 19.3 & 47.3 & 9.3 & 30.9 & 15.97 & 0.879 \\
    	$C=256$ & 18.7 & 46.5 & 8.7 & 30.5 & 27.98 & 1.096 \\
    	\bottomrule
    \end{tabular}
    }
    \vspace{-0.2cm}
\end{table} 

\label{sec:experiments_ablation}

%\vspace{-0.2cm}
\paragraph{Routing} We compare the proposed self-attention routing with previous routing methods including dynamic \cite{sabour2017dynamic}, EM \cite{hinton2018matrix}, and Set Transformer \cite{lee2019set} routing, as well as with a setup without any routing (i.e. learning a MLP to obtain existence probabilities). As dynamic and EM routing involve a computationally expensive iterative procedure and EM routing requires matrix capsules, we reduce the size of the network and fix the number of capsules to $C=64$ and the dimensionality of the primary and secondary capsule to $d_1=d_2=16$ to allow for a training with same batch size for all approaches.
%\MS{(this is an important point, it should be highlighted in introduction and abstract. From my point of view, the reason capsules are not that popular is that routing is hard to understand for many people. Your paper simplifies routing using self-attention. This work also shade a light on the comment by Hinton that capsules and attention/transformers has some relationship.)}. 
From the results shown in Table \ref{tab:ablation-routing}, we see that training with routing tends to outperform the respective baseline architectures without routing mechanisms. Among the evaluated methods, only the EM routing algorithm does not seem to be well suited for the targeted setup, as it suffers greatly from instability during training. In fact, all routing methods, including the proposed approach, tend to have some level of instability in this multimodal training paradigm, requiring the need for fine-tuning of learning rates. However, we find that our method, closely followed dynamic routing, suffer least from this instability and are able to achieve relatively strong performance in this experimental setup. 
%Overall, the proposed routing by self-attention outperforms previous routing algorithms, closely followed by dynamic routing which also achieved relatively strong performance in this experimental setup. 
One problem with iterative routing procedures, including dynamic routing, is that it becomes difficult to scale, mainly because of the larger memory footprint. Here, especially in the direct comparison with dynamic routing, the proposed method is able to achieve better results with fewer computational resources. 

% On the other hand, self-attention routing can easily increase the number of capsules by up to 4 times ($C=256$) on the same hardware.
\vspace{-0.3cm}
\paragraph{Number of Capsules} To show the ability of the proposed routing mechanism to scale, we also analyse how the number of capsules effects our proposed architecture. For these experiments, we maintain the capsule dimension of the original training setting with $d_1=32$ and $d_2=256$ while varying the number of capsules, $C=32,64,128,256$. As shown in Table \ref{tab:ablation-numcaps}, increasing the number of capsules generally leads to an improvement in performance. This can be seen as a indicator that a larger number of capsules allows the network to capture more object or concept representations. With the current dataset, we find that our models saturate at $C\geq 128$; when the number of capsules becomes larger, we find that there is a diminishing return on performance. Considering computational efficiency, it further shows that even for large numbers of capsules, the run-time is still below that of the iterative routing mechanisms.

\begin{table}[t]
    % \tablestyle{2pt}{1.05}
    \caption{Evaluation using fully connected and self-attention baselines. 
    \label{tab:ablation-arch}
    \vspace{-0.2cm}
    }
    \centering
    \resizebox{1\columnwidth}{!}{
    \begin{tabular}{@{}l|cccc|cccc@{}}
    	\toprule
    	\multicolumn{1}{c}{} & \multicolumn{4}{c}{YouCook2} & \multicolumn{4}{c}{MSR-VTT}  \\ 
    	\cmidrule(lr){2-5} \cmidrule(lr){6-9} 
    	Modalities  & R@1 & R@5  & R@10 & Med. R & R@1 & R@5 & R@10  & Med. R \\ 
    	\midrule
    	Fully Connected & 15.5 & 31.0 & 40.0 & 22 & 7.8 & 17.8 & 25.2 & 50 \\
    	Self-Attention & 13.8 & 29.3 & 36.2 & 34 & 8.9 & 22.3 & 30.1 & 41 \\
    	Ours & 19.3 & 37.8 & 47.3 & 13 & 9.3 & 21.4 & 30.9 & 37 \\
    	\bottomrule
    \end{tabular}
    }
    \vspace{-0.2cm}
\end{table} 

\vspace{-0.2cm}
\paragraph{Comparison with Fully Connected and Self-Attention Baselines}
%Additionally, we assess the performance of two baseline configurations, one with only a fully-connected layer  instead of capsule routing and one with a self-attention based layer, but without capsule routing. 
Since our main contribution is the proposal of a capsule-based framework for multimodal learning, we compare with other architecture baselines in Table \ref{tab:ablation-arch}. For a standard baseline, we have run an experiment which takes the input features and passes them through two fully connected layers (Fully Connected). It achieves lower performance than our proposed capsule network, showing that using capsules is valuable in learning multi-modal representations. We also compare
%We include an additional ablation to compare 
with self-attention without capsule structures. For this experiment, we apply a multi-head self-attention layer on the input features. We take the input features and group the activations into $N$ equal length vectors. Here, $N=128$ so that it is as similar to the number of capsules in our main experiments. These vectors are used as the sequence for a self-attention layer, which is followed by a fully-connected layer to obtain the final feature representation for each modality. 
We find that self-attention outperforms the fully connected baseline on MSR-VTT, but does not reach the performance of our proposed self-attention routing method.
%We find that self-attention without capsules under-performs capsules with our proposed self-attention routing method.

\begin{figure*}
    \centering
         
     \begin{subfigure}[t]{0.333\textwidth}
     \caption*{Ours}
     \end{subfigure}%
     \begin{subfigure}[t]{0.333\textwidth}
     \caption*{No routing}
     \end{subfigure}%
     \begin{subfigure}[t]{0.333\textwidth}
     \caption*{MIL NCE*}
     \end{subfigure}%
     \vspace{-4mm}
     
     \caption*{query: melt butter in the oven}
     \begin{subfigure}[t]{0.320\textwidth}
         \begin{subfigure}[t]{0.333\textwidth}
             \centering
            \includegraphics[width=\textwidth]{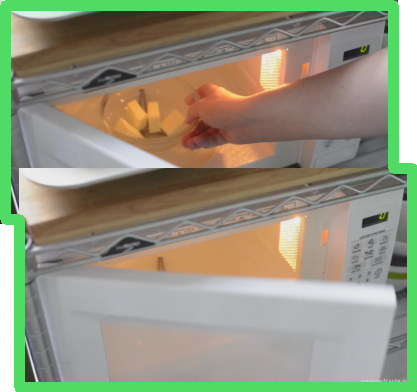}
         \end{subfigure}%
         \begin{subfigure}[t]{0.333\textwidth}
             \centering
             \includegraphics[width=\textwidth]{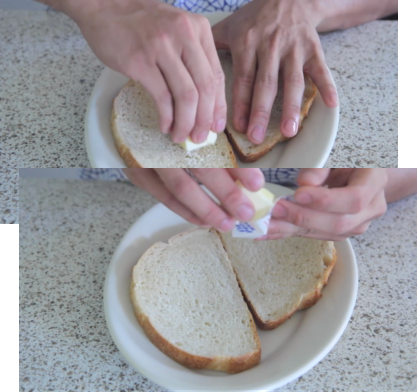}
         \end{subfigure}%
         \begin{subfigure}[t]{0.333\textwidth}
             \centering
             \includegraphics[width=\textwidth]{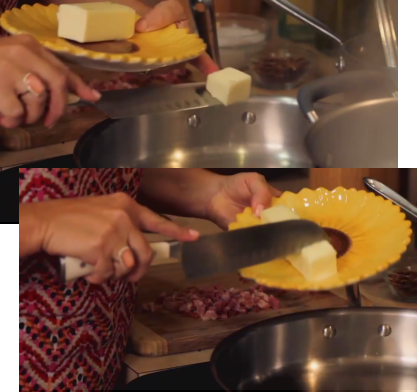}
         \end{subfigure}%
     \end{subfigure}%
     \hspace{0.019\textwidth}%
     \begin{subfigure}[t]{0.320\textwidth}
         \begin{subfigure}[t]{0.333\textwidth}
             \centering
            \includegraphics[width=\textwidth]{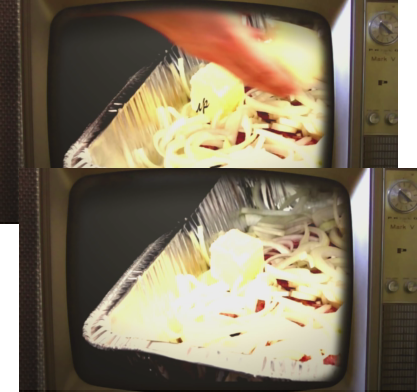}
         \end{subfigure}%
         \begin{subfigure}[t]{0.333\textwidth}
             \centering
             \includegraphics[width=\textwidth]{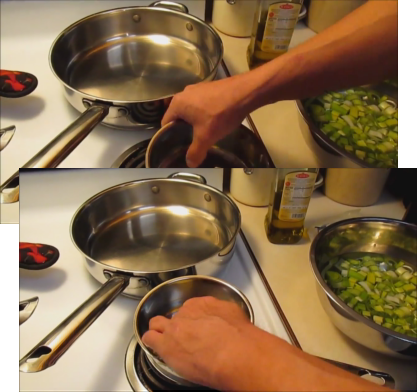}
         \end{subfigure}%
         \begin{subfigure}[t]{0.333\textwidth}
             \centering
             \includegraphics[width=\textwidth]{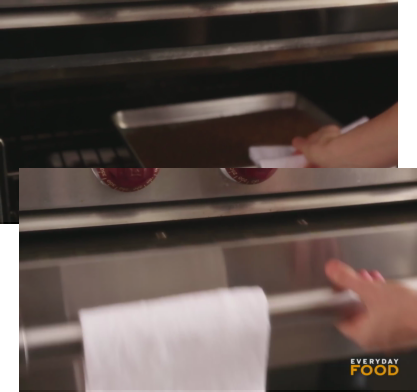}
         \end{subfigure}%
     \end{subfigure}%
     \hspace{0.019\textwidth}%
     \begin{subfigure}[t]{0.320\textwidth}
         \begin{subfigure}[t]{0.333\textwidth}
             \centering
            \includegraphics[width=\textwidth]{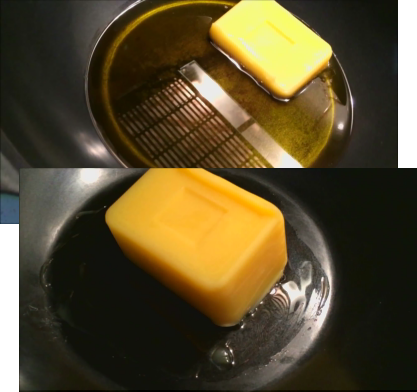}
         \end{subfigure}%
         \begin{subfigure}[t]{0.333\textwidth}
             \centering
             \includegraphics[width=\textwidth]{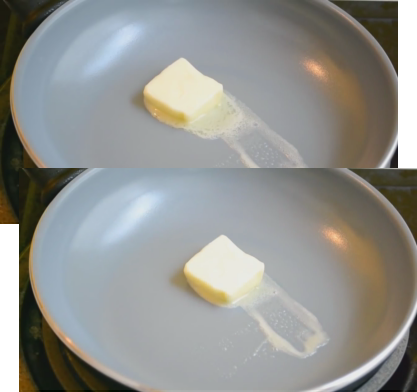}
         \end{subfigure}%
         \begin{subfigure}[t]{0.333\textwidth}
             \centering
             \includegraphics[width=\textwidth]{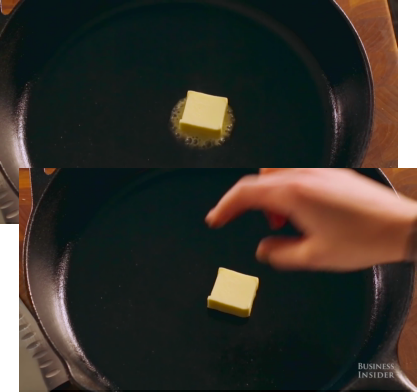}
         \end{subfigure}%
     \end{subfigure}
     
     \caption*{query: put three rings of ketchup and two rings of mustard on the bottom bun}
     \begin{subfigure}[t]{0.320\textwidth}
         \begin{subfigure}[t]{0.333\textwidth}
             \centering
            \includegraphics[width=\textwidth]{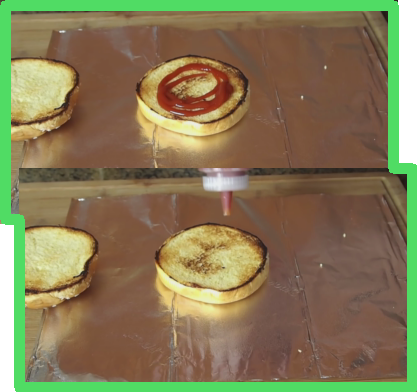}
         \end{subfigure}%
         \begin{subfigure}[t]{0.333\textwidth}
             \centering
             \includegraphics[width=\textwidth]{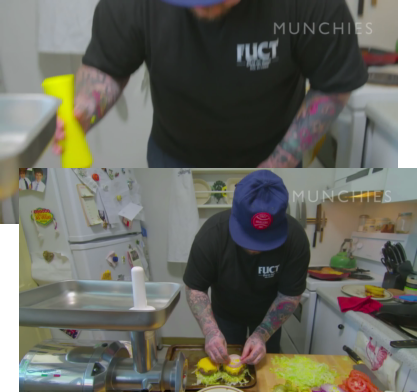}
         \end{subfigure}%
         \begin{subfigure}[t]{0.333\textwidth}
             \centering
             \includegraphics[width=\textwidth]{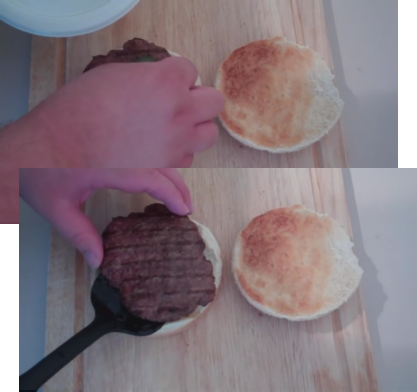}
         \end{subfigure}%
     \end{subfigure}%
     \hspace{0.019\textwidth}%
     \begin{subfigure}[t]{0.320\textwidth}
         \begin{subfigure}[t]{0.333\textwidth}
             \centering
            \includegraphics[width=\textwidth]{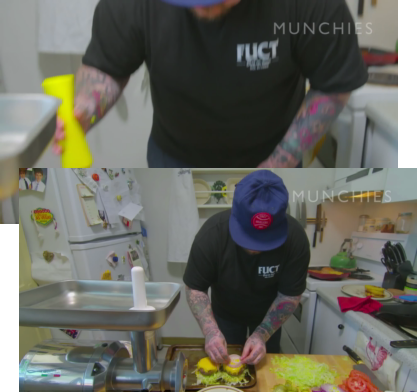}
         \end{subfigure}%
         \begin{subfigure}[t]{0.333\textwidth}
             \centering
             \includegraphics[width=\textwidth]{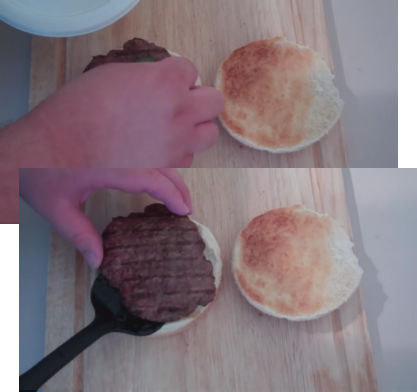}
         \end{subfigure}%
         \begin{subfigure}[t]{0.333\textwidth}
             \centering
             \includegraphics[width=\textwidth]{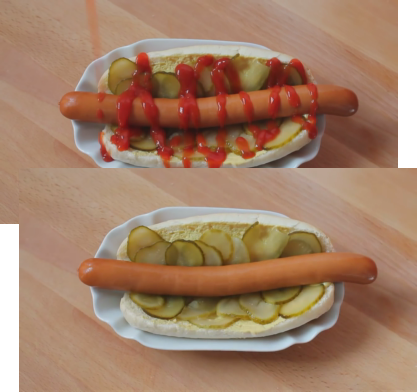}
         \end{subfigure}%
     \end{subfigure}%
     \hspace{0.019\textwidth}%
     \begin{subfigure}[t]{0.320\textwidth}
         \begin{subfigure}[t]{0.333\textwidth}
             \centering
            \includegraphics[width=\textwidth]{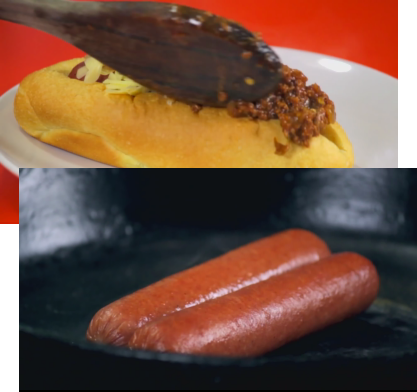}
         \end{subfigure}%
         \begin{subfigure}[t]{0.333\textwidth}
             \centering
             \includegraphics[width=\textwidth]{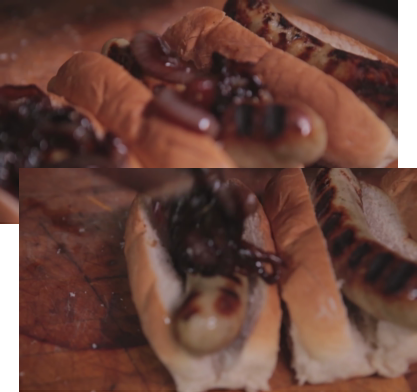}
         \end{subfigure}%
         \begin{subfigure}[t]{0.333\textwidth}
             \centering
             \includegraphics[width=\textwidth]{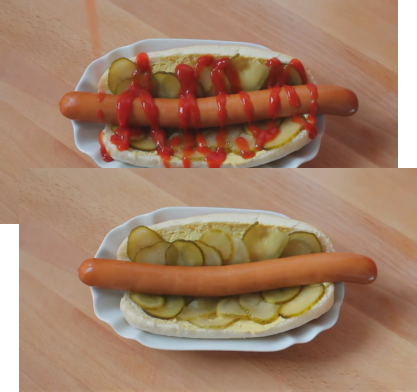}
         \end{subfigure}%
     \end{subfigure}

     \caption{Qualitative retrieval examples: top-3 zero-shot text-to-video retrieval results on the YouCook2 dataset for the proposed approach with self-attention based routing, our approach without the routing mechanism, and MIL-NCE* (* indicates that we used the same backbone as in our model). Correct video colored in green.}
    \label{fig:retrieval-main}
\end{figure*}
\begin{figure*}
    \centering
    \begin{subfigure}[t]{0.24\textwidth}
        \begin{subfigure}[t]{0.49\textwidth}
            \centering
            \includegraphics[width=\textwidth]{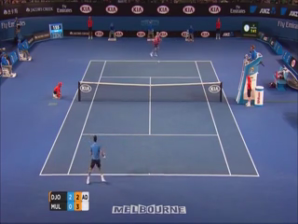}
         \end{subfigure}%
         \hspace{0.01\textwidth}%
         \begin{subfigure}[t]{0.49\textwidth}
             \centering
            \includegraphics[width=\textwidth]{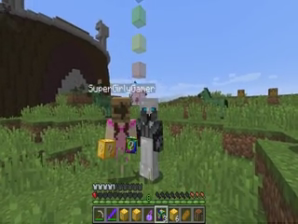}
         \end{subfigure}
         \begin{subfigure}[t]{0.49\textwidth}
            \centering
            \includegraphics[width=\textwidth]{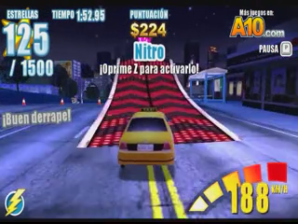}
         \end{subfigure}%
         \hspace{0.01\textwidth}%
         \begin{subfigure}[t]{0.49\textwidth}
             \centering
            \includegraphics[width=\textwidth]{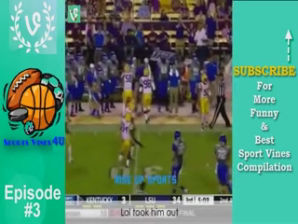}
         \end{subfigure}
         \caption*{\footnotesize{\#81: games}}
     \end{subfigure}%
     \hspace{0.01\textwidth}%
     \begin{subfigure}[t]{0.24\textwidth}
        \begin{subfigure}[t]{0.49\textwidth}
            \centering
            \includegraphics[width=\textwidth]{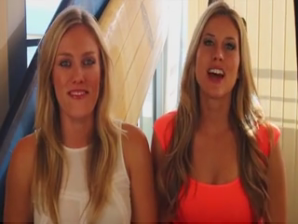}
         \end{subfigure}%
         \hspace{0.01\textwidth}%
         \begin{subfigure}[t]{0.49\textwidth}
             \centering
            \includegraphics[width=\textwidth]{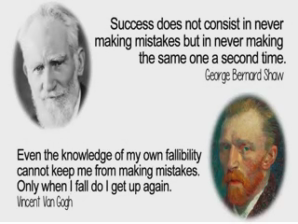}
         \end{subfigure}
         \begin{subfigure}[t]{0.49\textwidth}
            \centering
            \includegraphics[width=\textwidth]{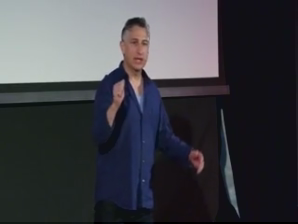}
         \end{subfigure}%
         \hspace{0.01\textwidth}%
         \begin{subfigure}[t]{0.49\textwidth}
             \centering
            \includegraphics[width=\textwidth]{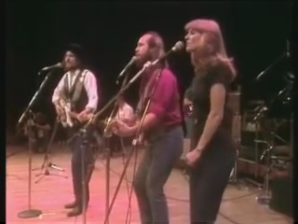}
         \end{subfigure}
         \caption*{\footnotesize{\#98: people}}
     \end{subfigure}%
     \hspace{0.01\textwidth}%
     \begin{subfigure}[t]{0.24\textwidth}
        \begin{subfigure}[t]{0.49\textwidth}
            \centering
            \includegraphics[width=\textwidth]{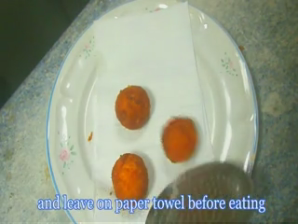}
         \end{subfigure}%
         \hspace{0.01\textwidth}%
         \begin{subfigure}[t]{0.49\textwidth}
             \centering
            \includegraphics[width=\textwidth]{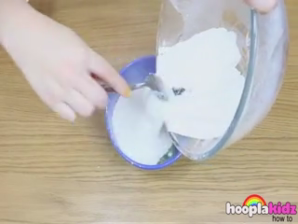}
         \end{subfigure}
         \begin{subfigure}[t]{0.49\textwidth}
            \centering
            \includegraphics[width=\textwidth]{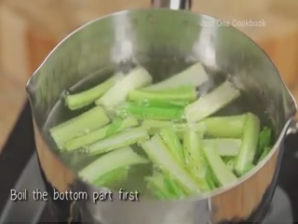}
         \end{subfigure}%
         \hspace{0.01\textwidth}%
         \begin{subfigure}[t]{0.49\textwidth}
             \centering
            \includegraphics[width=\textwidth]{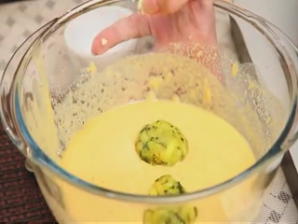}
         \end{subfigure}
         \caption*{\footnotesize{\#49: pots/bowls}}
     \end{subfigure}%
     \hspace{0.01\textwidth}%
     \begin{subfigure}[t]{0.24\textwidth}
        \begin{subfigure}[t]{0.49\textwidth}
            \centering
            \includegraphics[width=\textwidth]{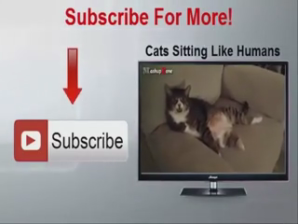}
         \end{subfigure}%
         \hspace{0.01\textwidth}%
         \begin{subfigure}[t]{0.49\textwidth}
             \centering
            \includegraphics[width=\textwidth]{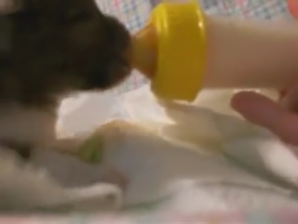}
         \end{subfigure}
         \begin{subfigure}[t]{0.49\textwidth}
            \centering
            \includegraphics[width=\textwidth]{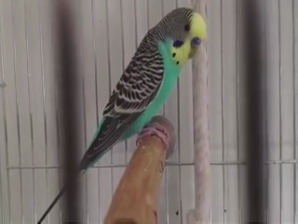}
         \end{subfigure}%
         \hspace{0.01\textwidth}%
         \begin{subfigure}[t]{0.49\textwidth}
             \centering
            \includegraphics[width=\textwidth]{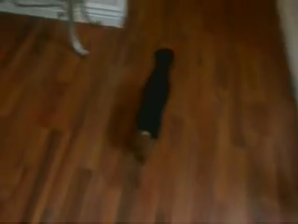}
         \end{subfigure}
         \caption*{\footnotesize{\#28: pets/animals}}
     \end{subfigure}
     
     \begin{subfigure}[t]{0.24\textwidth}
        \begin{subfigure}[t]{0.49\textwidth}
            \centering
            \includegraphics[width=\textwidth]{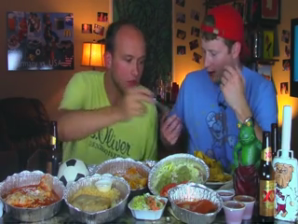}
         \end{subfigure}%
         \hspace{0.01\textwidth}%
         \begin{subfigure}[t]{0.49\textwidth}
             \centering
            \includegraphics[width=\textwidth]{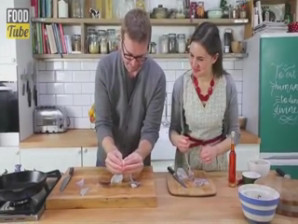}
         \end{subfigure}
         \begin{subfigure}[t]{0.49\textwidth}
            \centering
            \includegraphics[width=\textwidth]{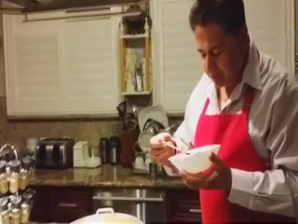}
         \end{subfigure}%
         \hspace{0.01\textwidth}%
         \begin{subfigure}[t]{0.49\textwidth}
             \centering
            \includegraphics[width=\textwidth]{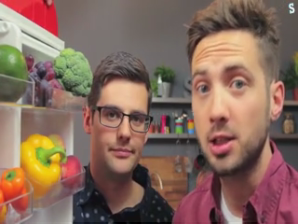}
         \end{subfigure}
         \caption*{\footnotesize{\#60: cooking}}
     \end{subfigure}%
     \hspace{0.01\textwidth}%
     \begin{subfigure}[t]{0.24\textwidth}
        \begin{subfigure}[t]{0.49\textwidth}
            \centering
            \includegraphics[width=\textwidth]{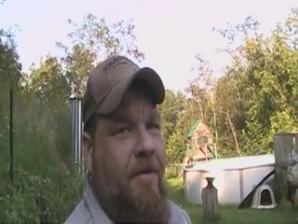}
         \end{subfigure}%
         \hspace{0.01\textwidth}%
         \begin{subfigure}[t]{0.49\textwidth}
             \centering
            \includegraphics[width=\textwidth]{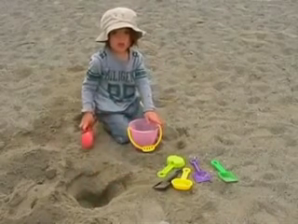}
         \end{subfigure}
         \begin{subfigure}[t]{0.49\textwidth}
            \centering
            \includegraphics[width=\textwidth]{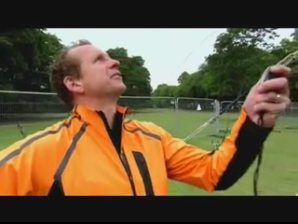}
         \end{subfigure}%
         \hspace{0.01\textwidth}%
         \begin{subfigure}[t]{0.49\textwidth}
             \centering
            \includegraphics[width=\textwidth]{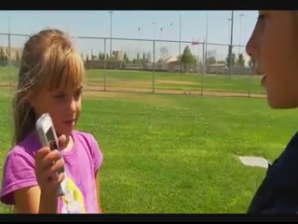}
         \end{subfigure}
         \caption*{\footnotesize{\#110: outdoor activity}}
     \end{subfigure}%
     \hspace{0.01\textwidth}%
     \begin{subfigure}[t]{0.24\textwidth}
        \begin{subfigure}[t]{0.49\textwidth}
            \centering
            \includegraphics[width=\textwidth]{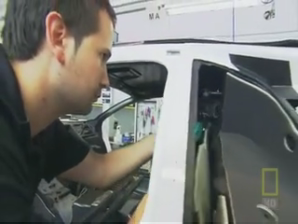}
         \end{subfigure}%
         \hspace{0.01\textwidth}%
         \begin{subfigure}[t]{0.49\textwidth}
             \centering
            \includegraphics[width=\textwidth]{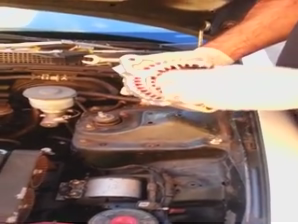}
         \end{subfigure}
         \begin{subfigure}[t]{0.49\textwidth}
            \centering
            \includegraphics[width=\textwidth]{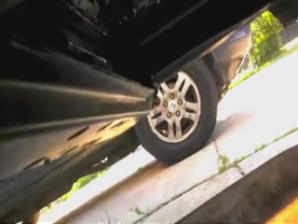}
         \end{subfigure}%
         \hspace{0.01\textwidth}%
         \begin{subfigure}[t]{0.49\textwidth}
             \centering
            \includegraphics[width=\textwidth]{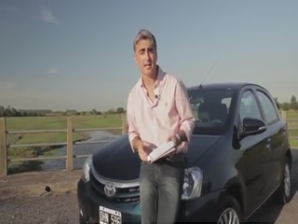}
         \end{subfigure}
         \caption*{\footnotesize{\#85: cars}}
     \end{subfigure}%
     \hspace{0.01\textwidth}%
     \begin{subfigure}[t]{0.24\textwidth}
        \begin{subfigure}[t]{0.49\textwidth}
            \centering
            \includegraphics[width=\textwidth]{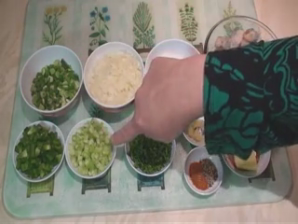}
         \end{subfigure}%
         \hspace{0.01\textwidth}%
         \begin{subfigure}[t]{0.49\textwidth}
             \centering
            \includegraphics[width=\textwidth]{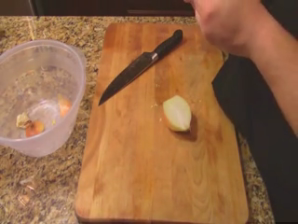}
         \end{subfigure}
         \begin{subfigure}[t]{0.49\textwidth}
            \centering
            \includegraphics[width=\textwidth]{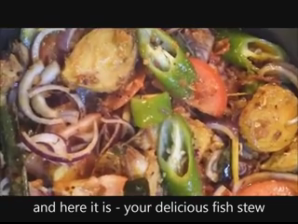}
         \end{subfigure}%
         \hspace{0.01\textwidth}%
         \begin{subfigure}[t]{0.49\textwidth}
             \centering
            \includegraphics[width=\textwidth]{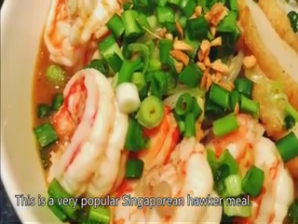}
         \end{subfigure}
         \caption*{\footnotesize{\#123: vegetables}}
     \end{subfigure}%
     \caption{Top-4 videos with the highest activation for the particular capsule on the MSR-VTT dataset. Labels: \#number of capsule: assumed learned ``concept". }
    \label{fig:capsule-acts}
    \vspace{-3mm}
\end{figure*}
%\MS{This is nice, these are video capsules, how about text and audio capsules?}

%\vspace{-0.2cm}
\subsection{Qualitative Analysis}
\label{sec:experiments_qualitative}

In our final set of evaluations, we attempt to understand what the proposed architecture is able to learn by analysing a set of qualitative retrieval examples as well as studying how individual capsule activations effect the final feature representation. We include additional qualitative results in the appendix.

\vspace{-2mm}
\paragraph{Retrieval Results}
We present retrieval results for three models - our self-attention based routing method, our approach without routing, and MIL-NCE - in Figure \ref{fig:retrieval-main}. Each column consists of the top-3 predictions for the given text query. Generally, routing achieves strong performance and retrieves visually varied videos; on the other hand, MIL-NCE tends to focus on specific objects or low-level visual cues leading to visually similar retrievals. In the first example, MIL-NCE retrieves videos of ``melt butter", but the butter is melted in a pan and not an ``oven". Notably, our approach successfully handles the extremely specific query ``Put three rings of ketchup and two rings of mustard on the bottom bun" as shown in the second row. %Additional qualitative results are presented in Appendix \ref{sec:app:qual}.

\vspace{-2mm}
\paragraph{What Individual Capsules Learn} 
To further understand the entities or objects that are modeled, we examine the capsules' activations $\hat{p}_i^m$ (Equation \ref{eq:acts}) and show samples that have a high activation for a specific capsule. Ideally, if two samples have a high activation for the same capsule, then the entity that it represents should be present within both inputs. In the MSRVTT dataset we select the videos which lead to high activations for various capsules; we observe that different capsules model semantically distinct concepts as seen in Figure \ref{fig:capsule-acts}. The capsules learn to represent a wide range of entities: from general concepts like ``games", ``cooking", and ``outdoor activities", to specific objects like ``vegetables" and ``cars". We find that this behaviour is consistent across various datasets.  %In Appendix \ref{sec:app:qual} we include additional examples and show that these concepts are consistent across different datasets. 

%\subsection{Limitations}
%\label{sec:experiments_limitations}

%Considering the limitations of the proposed approach, we found that the training of the system with routing methods (see Tab. \ref{tab:ablation-routing}) tends to \hkc{collapse} more often than its fully connected of self-attention baseline, with EM routing being the most unstable routing mechanism. We attribute this to \hkc{ ... do we have an idea ?}. We also observe our approach seems to be the most stable one leading to the least collapses among all routing techniques.
%\hkc{Perhaps add learning rate discussion ... }
%We also notice that the initial learning rate training stability tends to vary largely 

%We further observe that systems trained in a multimodal setup might be more in general is more sensitive to domain shifts as compared to single modality tasks like image classification. We attribute that to the fact that dealing with multiple modalities obviously also extends the possibility of domain shifts, as it can be enough for one domain to deviate from the original training data to actually reduce the performance of the model. This can be seen in the comparison of the performance of different modality combinations of MSRVTT. But it also shows that in this case, better performance can be reach by simply dropping the misaligned domain.

\section{Conclusion}

In this work, we proposed a novel multimodal capsule network that learns to model various entities within given modalities and maps them to a joint embedding space. To learn from a large amount of noisy video data, we present a scalable self-attention based capsule routing mechanism, which we show outperforms previous routing methods on this task. Furthermore, we find that the capsules are able to learn representations of various concepts and objects within each modality. Our comprehensive experimental evaluation demonstrates the effectiveness of our approach on two downstream zero-shot tasks on four datasets.

%%%%%%%%% REFERENCES
{\small
\bibliographystyle{ieee_fullname}
\bibliography{egbib}
}

\appendix
\section{Additional Ablations}
\label{sec:app:ablations}

\paragraph{Shared weights}
For our proposed architecture, we share weights across the various modalities after the initial capsules are extracted. Not only does this reduce the number of learned parameters for the network, but we find that it leads to learning improved representations. We present results in Table \ref{tab:ablation-shared}. Generally, for data more closely related to the training data (for example, evaluating on YouCook2) the use of shared weights leads to improved performance.

\begin{table}
    % \tablestyle{2pt}{1.05}
    \caption{Evaluation using shared weights 
    \label{tab:ablation-shared}
    }
    \centering
    
    \resizebox{1\columnwidth}{!}{
    \begin{tabular}{@{}l|cccc|cccc@{}}
    	\toprule
    	\multicolumn{1}{c}{} & \multicolumn{4}{c}{YouCook2} & \multicolumn{4}{c}{MSR-VTT}  \\ 
    	\cmidrule(lr){2-5} \cmidrule(lr){6-9} 
    	Modalities  & R@1 & R@5  & R@10 & Med. R & R@1 & R@5 & R@10  & Med. R \\ 
    	\midrule
        % R@1: 0.1681 - R@5: 0.3540 - R@10: 0.4457 - Median R: 15.0
        % R@1: 0.0950 - R@5: 0.2280 - R@10: 0.3030 - Median R: 30.5
    	Not Shared & 16.8 & 35.4 & 44.6 & 15 & 9.5 & 22.8 & 30.3 & 30.5   \\
    	Shared & 19.3 & 37.8 & 47.3 & 13 & 9.6 & 22.6 & 32.0 & 32 \\
    	\bottomrule
    \end{tabular}
    }
    % \vspace{-0.25cm}
\end{table}

\section{Self-attention Architectural Details} \label{app:arch}
For our self-attention routing procedure we first use linear projections to generate the query-key-value. Given that there are $C$ input capsules and the output capsules have dimension $d_2$, the query, key, and value matrices have shape $Q,K,V\in \mathbb{R}^{C\times d_2}$. The output of the multi-head self-attention operation,
\begin{equation}
    V'=\text{softmax}\left(\frac{QK^T}{\sqrt{d_2}}\right)V,
\end{equation}
is a matrix of the same dimension. We then apply normalization across the columns (i.e. capsule feature dimension) as well as two fully connected linear layers, and dropout, with hidden dimension 1024 and output dimension $d_2$. A residual connection from $V'$ to the output capsule features, followed by normalization across the capsule feature dimensions.

\section{Additional Qualitative Results}
\label{sec:app:qual}
\begin{figure*}
    \centering
    \begin{subfigure}[t]{0.25\textwidth}
    \centering
    \stackinset{c}{0pt}{c}{0pt}{\Centerstack{
    \footnotesize ``add oil vinegar lemon \\
    \footnotesize juice and garlic to a bowl"
    }}
    {\includegraphics[width=\textwidth]{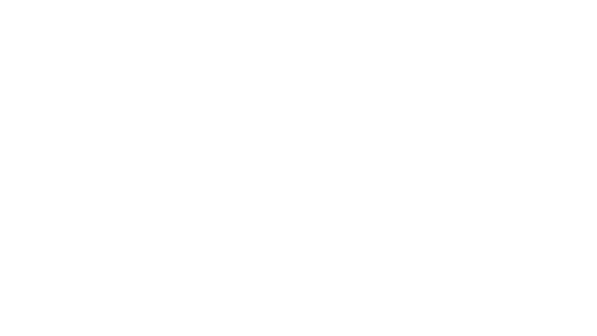}}
     \end{subfigure}
     \begin{subfigure}[t]{0.23\textwidth}
         \centering
        \includegraphics[width=\textwidth]{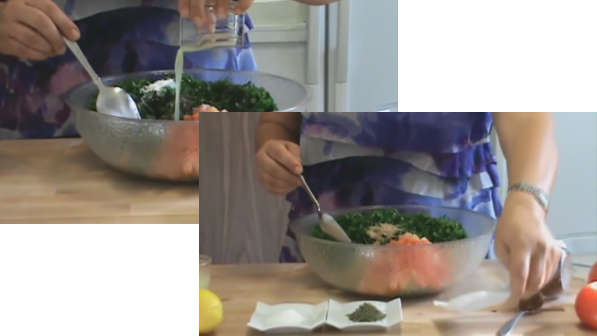}
     \end{subfigure}
     \begin{subfigure}[t]{0.23\textwidth}
         \centering
         \includegraphics[width=\textwidth]{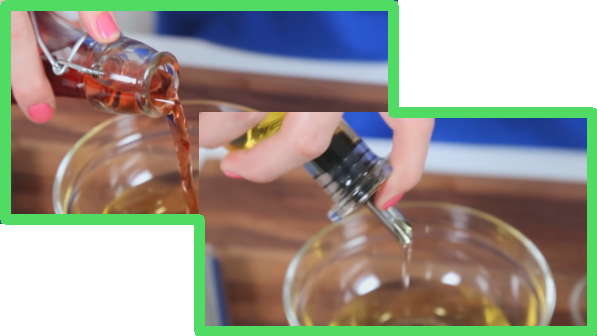}
     \end{subfigure}
     \begin{subfigure}[t]{0.23\textwidth}
         \centering
         \includegraphics[width=\textwidth]{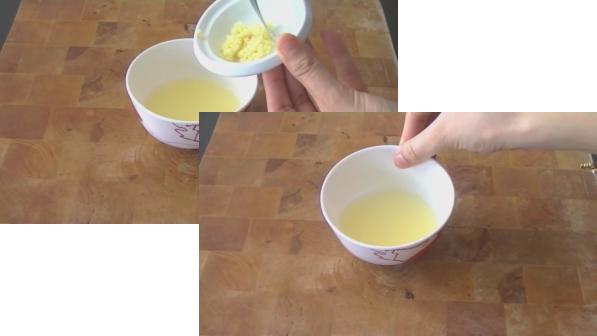}
     \end{subfigure}
     
     \begin{subfigure}[t]{0.25\textwidth}
    \centering
    \stackinset{c}{0pt}{c}{0pt}{\Centerstack{
    \footnotesize ``chop and add fresh basil \\
    \footnotesize  and stir"
    }}
    {\includegraphics[width=\textwidth]{Figures/text_to_video/blank.png}}
     \end{subfigure}
     \begin{subfigure}[t]{0.23\textwidth}
         \centering
        \includegraphics[width=\textwidth]{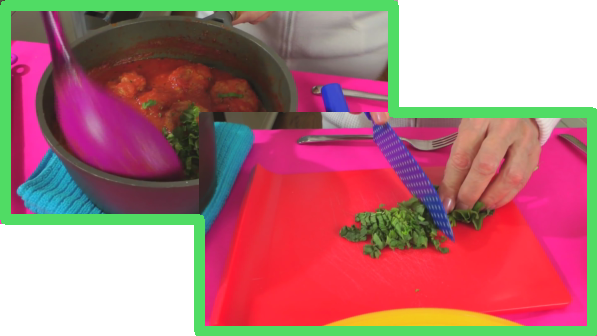}
     \end{subfigure}
     \begin{subfigure}[t]{0.23\textwidth}
         \centering
         \includegraphics[width=\textwidth]{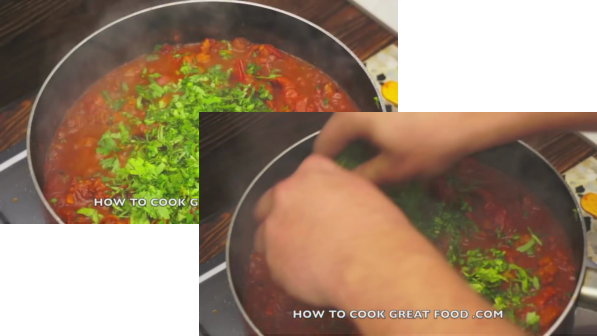}
     \end{subfigure}
     \begin{subfigure}[t]{0.23\textwidth}
         \centering
         \includegraphics[width=\textwidth]{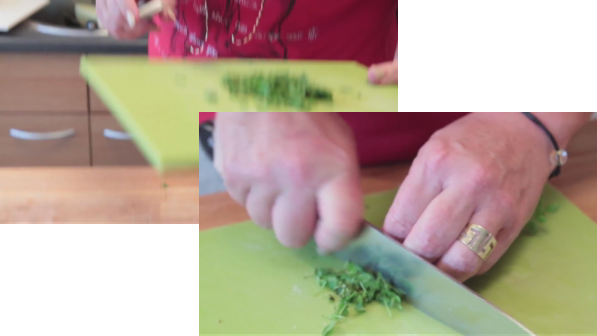}
     \end{subfigure}
     
     \begin{subfigure}[t]{0.25\textwidth}
    \centering
    \stackinset{c}{0pt}{c}{0pt}{\Centerstack{
    \footnotesize ``mix flour baking powder \\
    \footnotesize and salt"
    }}
    {\includegraphics[width=\textwidth]{Figures/text_to_video/blank.png}}
     \end{subfigure}
     \begin{subfigure}[t]{0.23\textwidth}
         \centering
        \includegraphics[width=\textwidth]{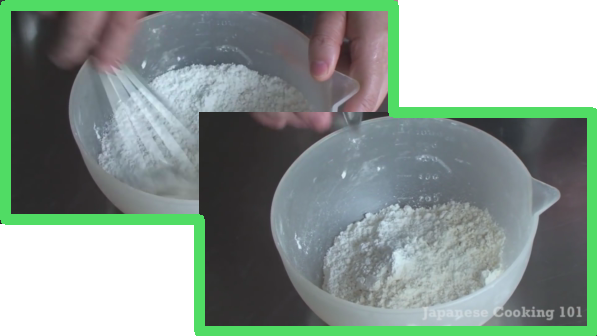}
     \end{subfigure}
     \begin{subfigure}[t]{0.23\textwidth}
         \centering
         \includegraphics[width=\textwidth]{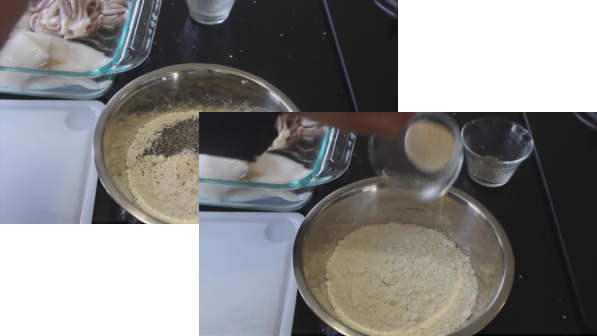}
     \end{subfigure}
     \begin{subfigure}[t]{0.23\textwidth}
         \centering
         \includegraphics[width=\textwidth]{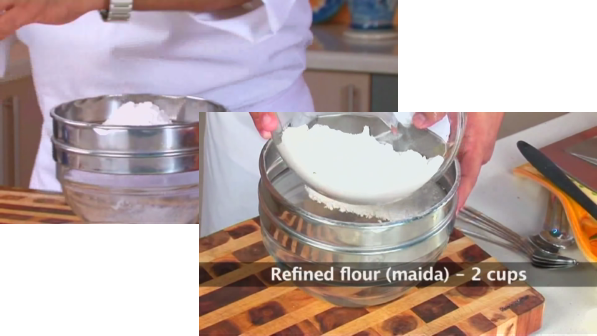}
     \end{subfigure}
     
     \begin{subfigure}[t]{0.25\textwidth}
    \centering
    \stackinset{c}{0pt}{c}{0pt}{\Centerstack{
    \footnotesize ``marinate the ribs well  \\
    \footnotesize  with the prepared sauce \\
    \footnotesize in a bucket overnight"
    }}
    {\includegraphics[width=\textwidth]{Figures/text_to_video/blank.png}}
     \end{subfigure}
     \begin{subfigure}[t]{0.23\textwidth}
         \centering
        \includegraphics[width=\textwidth]{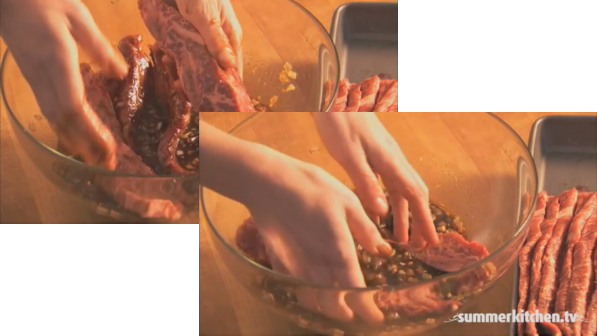}
     \end{subfigure}
     \begin{subfigure}[t]{0.23\textwidth}
         \centering
         \includegraphics[width=\textwidth]{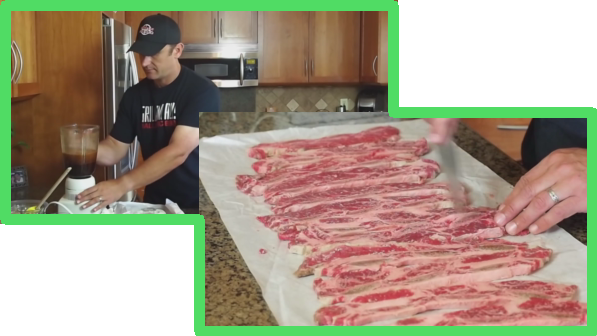}
     \end{subfigure}
     \begin{subfigure}[t]{0.23\textwidth}
         \centering
         \includegraphics[width=\textwidth]{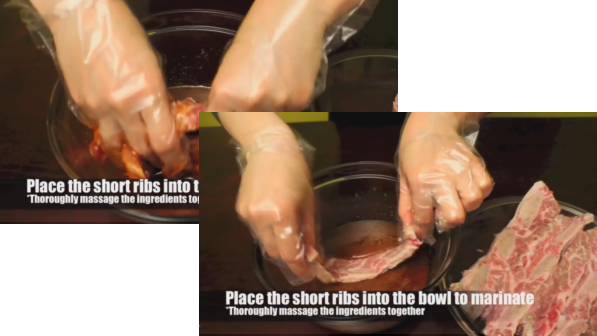}
     \end{subfigure}
     
     \caption{Qualitative evaluation: examples of top-3 zero-shot text-to-video retrieval results on the YouCook2 dataset. Correct video colored in green.}
    \label{fig:retrieval}
\end{figure*}

\paragraph{Retrieval Quality}
In the case of retrieval, we show three text queries together with their three closest video representations in Figure \ref{fig:retrieval}. It becomes clear that all video representations show a close match for the described scene. Additionally, one has to remark that the retrieved video examples for each query do show sufficient variance with respect to color, view point, and other low-level cues. This can be seen as an indicator that the learned clustering is based on some high-level common concepts rather than on the pure co-occurrence of low-level feature representations.

\begin{figure*}
    \centering
         
     \begin{subfigure}[t]{0.333\textwidth}
     \caption*{Ours}
     \end{subfigure}%
     \begin{subfigure}[t]{0.333\textwidth}
     \caption*{No routing}
     \end{subfigure}%
     \begin{subfigure}[t]{0.333\textwidth}
     \caption*{MIL NCE*}
     \end{subfigure}%
     \vspace{-2mm}
    
    \caption*{query: put three rings of ketchup and two rings of mustard on the bottom bun}
     \begin{subfigure}[t]{0.320\textwidth}
         \begin{subfigure}[t]{0.333\textwidth}
             \centering
            \includegraphics[width=\textwidth]{Figures/supplement_retrieval/408_ours_0.png}
         \end{subfigure}%
         \begin{subfigure}[t]{0.333\textwidth}
             \centering
             \includegraphics[width=\textwidth]{Figures/supplement_retrieval/408_ours_1.png}
         \end{subfigure}%
         \begin{subfigure}[t]{0.333\textwidth}
             \centering
             \includegraphics[width=\textwidth]{Figures/supplement_retrieval/408_ours_2.png}
         \end{subfigure}%
     \end{subfigure}%
     \hspace{0.019\textwidth}%
     \begin{subfigure}[t]{0.320\textwidth}
         \begin{subfigure}[t]{0.333\textwidth}
             \centering
            \includegraphics[width=\textwidth]{Figures/supplement_retrieval/408_noroute_0.png}
         \end{subfigure}%
         \begin{subfigure}[t]{0.333\textwidth}
             \centering
             \includegraphics[width=\textwidth]{Figures/supplement_retrieval/408_noroute_1.png}
         \end{subfigure}%
         \begin{subfigure}[t]{0.333\textwidth}
             \centering
             \includegraphics[width=\textwidth]{Figures/supplement_retrieval/408_noroute_2.png}
         \end{subfigure}%
     \end{subfigure}%
     \hspace{0.019\textwidth}%
     \begin{subfigure}[t]{0.320\textwidth}
         \begin{subfigure}[t]{0.333\textwidth}
             \centering
            \includegraphics[width=\textwidth]{Figures/supplement_retrieval/408_milnce_0.png}
         \end{subfigure}%
         \begin{subfigure}[t]{0.333\textwidth}
             \centering
             \includegraphics[width=\textwidth]{Figures/supplement_retrieval/408_milnce_1.png}
         \end{subfigure}%
         \begin{subfigure}[t]{0.333\textwidth}
             \centering
             \includegraphics[width=\textwidth]{Figures/supplement_retrieval/408_milnce_2.png}
         \end{subfigure}%
     \end{subfigure}
    
     \caption*{query: melt butter in the oven}
     \begin{subfigure}[t]{0.320\textwidth}
         \begin{subfigure}[t]{0.333\textwidth}
             \centering
            \includegraphics[width=\textwidth]{Figures/supplement_retrieval/138_ours_0.png}
         \end{subfigure}%
         \begin{subfigure}[t]{0.333\textwidth}
             \centering
             \includegraphics[width=\textwidth]{Figures/supplement_retrieval/138_ours_1.png}
         \end{subfigure}%
         \begin{subfigure}[t]{0.333\textwidth}
             \centering
             \includegraphics[width=\textwidth]{Figures/supplement_retrieval/138_ours_2.png}
         \end{subfigure}%
     \end{subfigure}%
     \hspace{0.019\textwidth}%
     \begin{subfigure}[t]{0.320\textwidth}
         \begin{subfigure}[t]{0.333\textwidth}
             \centering
            \includegraphics[width=\textwidth]{Figures/supplement_retrieval/138_noroute_0.png}
         \end{subfigure}%
         \begin{subfigure}[t]{0.333\textwidth}
             \centering
             \includegraphics[width=\textwidth]{Figures/supplement_retrieval/138_noroute_1.png}
         \end{subfigure}%
         \begin{subfigure}[t]{0.333\textwidth}
             \centering
             \includegraphics[width=\textwidth]{Figures/supplement_retrieval/138_noroute_2.png}
         \end{subfigure}%
     \end{subfigure}%
     \hspace{0.019\textwidth}%
     \begin{subfigure}[t]{0.320\textwidth}
         \begin{subfigure}[t]{0.333\textwidth}
             \centering
            \includegraphics[width=\textwidth]{Figures/supplement_retrieval/138_milnce_0.png}
         \end{subfigure}%
         \begin{subfigure}[t]{0.333\textwidth}
             \centering
             \includegraphics[width=\textwidth]{Figures/supplement_retrieval/138_milnce_1.png}
         \end{subfigure}%
         \begin{subfigure}[t]{0.333\textwidth}
             \centering
             \includegraphics[width=\textwidth]{Figures/supplement_retrieval/138_milnce_2.png}
         \end{subfigure}%
     \end{subfigure}
     
     \caption*{query: pour the egg mixture over the bacon}
     \begin{subfigure}[t]{0.320\textwidth}
         \begin{subfigure}[t]{0.333\textwidth}
             \centering
            \includegraphics[width=\textwidth]{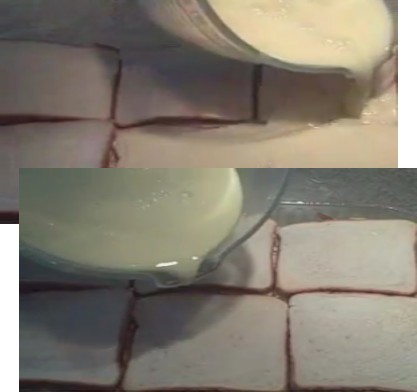}
         \end{subfigure}%
         \begin{subfigure}[t]{0.333\textwidth}
             \centering
             \includegraphics[width=\textwidth]{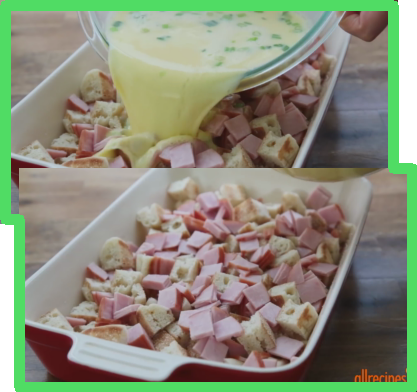}
         \end{subfigure}%
         \begin{subfigure}[t]{0.333\textwidth}
             \centering
             \includegraphics[width=\textwidth]{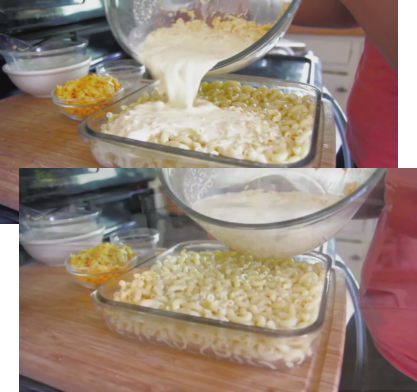}
         \end{subfigure}%
     \end{subfigure}%
     \hspace{0.019\textwidth}%
     \begin{subfigure}[t]{0.320\textwidth}
         \begin{subfigure}[t]{0.333\textwidth}
             \centering
            \includegraphics[width=\textwidth]{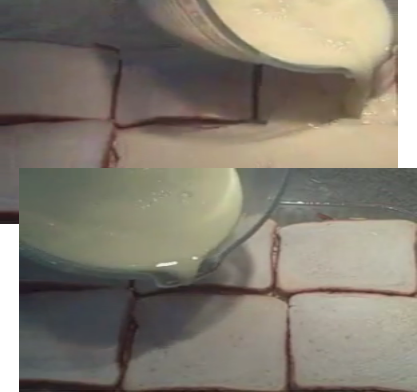}
         \end{subfigure}%
         \begin{subfigure}[t]{0.333\textwidth}
             \centering
             \includegraphics[width=\textwidth]{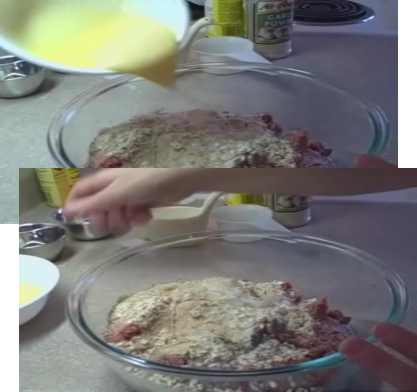}
         \end{subfigure}%
         \begin{subfigure}[t]{0.333\textwidth}
             \centering
             \includegraphics[width=\textwidth]{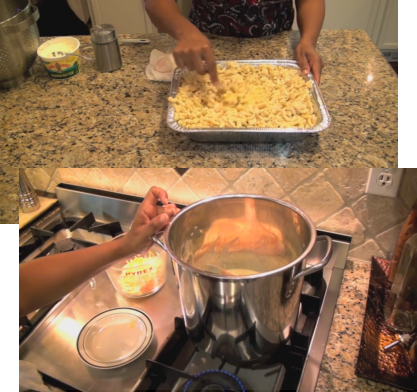}
         \end{subfigure}%
     \end{subfigure}%
     \hspace{0.019\textwidth}%
     \begin{subfigure}[t]{0.320\textwidth}
         \begin{subfigure}[t]{0.333\textwidth}
             \centering
            \includegraphics[width=\textwidth]{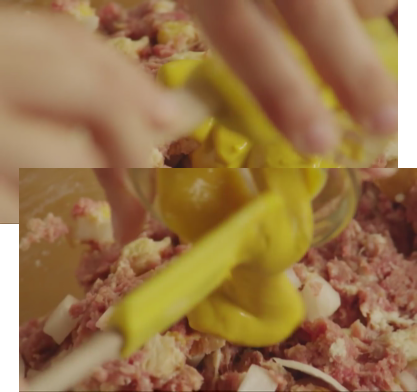}
         \end{subfigure}%
         \begin{subfigure}[t]{0.333\textwidth}
             \centering
             \includegraphics[width=\textwidth]{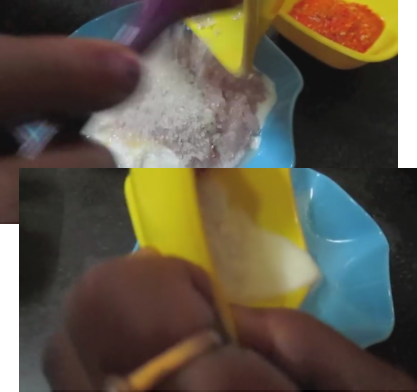}
         \end{subfigure}%
         \begin{subfigure}[t]{0.333\textwidth}
             \centering
             \includegraphics[width=\textwidth]{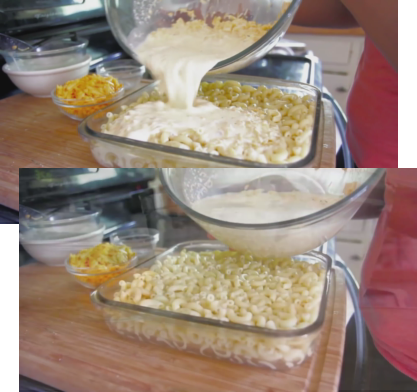}
         \end{subfigure}%
     \end{subfigure}
     
     \caption*{query: spread butter and maple syrup on the pancakes}
     \begin{subfigure}[t]{0.320\textwidth}
         \begin{subfigure}[t]{0.333\textwidth}
             \centering
            \includegraphics[width=\textwidth]{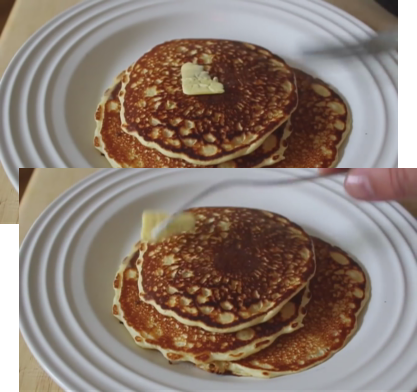}
         \end{subfigure}%
         \begin{subfigure}[t]{0.333\textwidth}
             \centering
             \includegraphics[width=\textwidth]{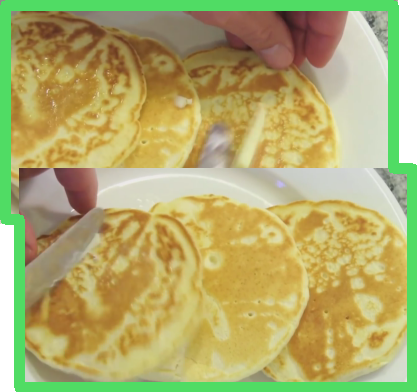}
         \end{subfigure}%
         \begin{subfigure}[t]{0.333\textwidth}
             \centering
             \includegraphics[width=\textwidth]{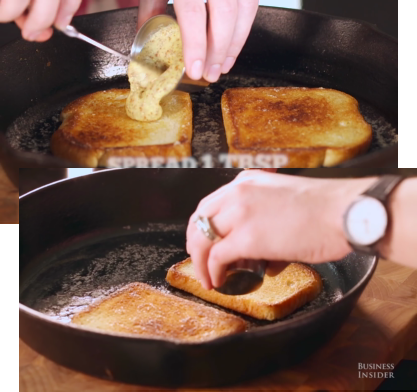}
         \end{subfigure}%
     \end{subfigure}%
     \hspace{0.019\textwidth}%
     \begin{subfigure}[t]{0.320\textwidth}
         \begin{subfigure}[t]{0.333\textwidth}
             \centering
            \includegraphics[width=\textwidth]{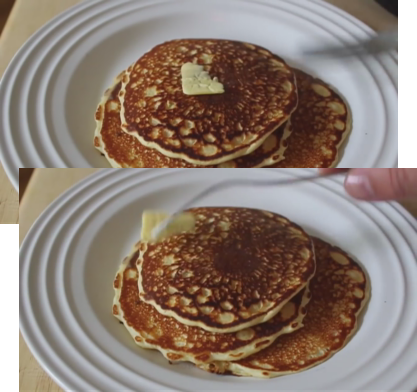}
         \end{subfigure}%
         \begin{subfigure}[t]{0.333\textwidth}
             \centering
             \includegraphics[width=\textwidth]{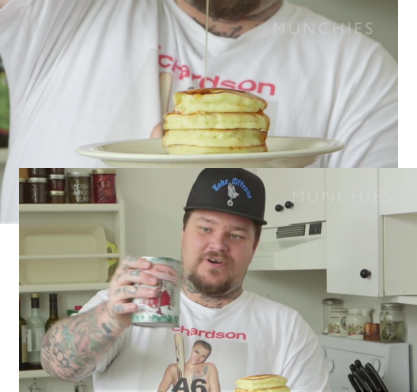}
         \end{subfigure}%
         \begin{subfigure}[t]{0.333\textwidth}
             \centering
             \includegraphics[width=\textwidth]{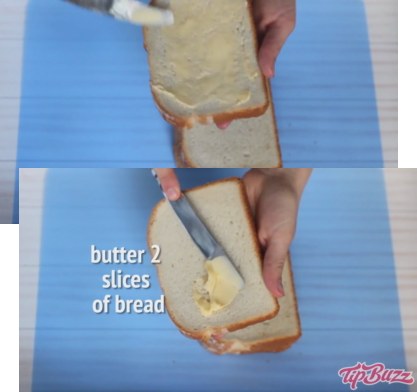}
         \end{subfigure}%
     \end{subfigure}%
     \hspace{0.019\textwidth}%
     \begin{subfigure}[t]{0.320\textwidth}
         \begin{subfigure}[t]{0.333\textwidth}
             \centering
            \includegraphics[width=\textwidth]{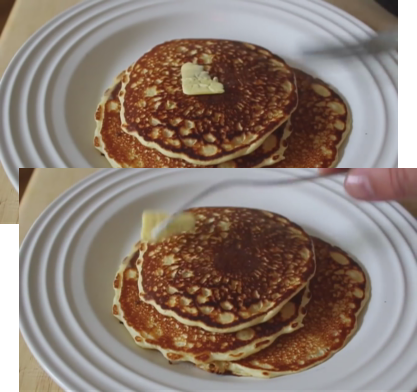}
         \end{subfigure}%
         \begin{subfigure}[t]{0.333\textwidth}
             \centering
             \includegraphics[width=\textwidth]{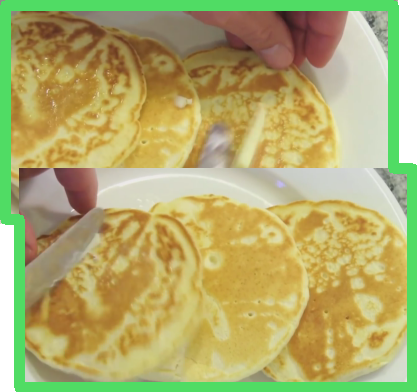}
         \end{subfigure}%
         \begin{subfigure}[t]{0.333\textwidth}
             \centering
             \includegraphics[width=\textwidth]{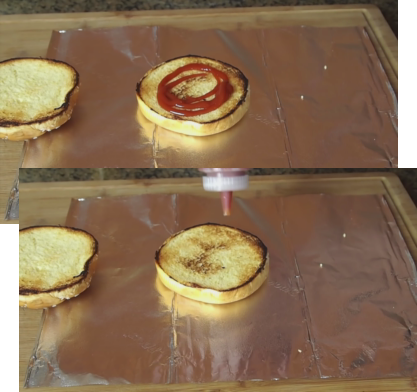}
         \end{subfigure}%
     \end{subfigure}
     
     \caption*{query: add lemon juice white wine and the mussels to the pot}
     \begin{subfigure}[t]{0.320\textwidth}
         \begin{subfigure}[t]{0.333\textwidth}
             \centering
            \includegraphics[width=\textwidth]{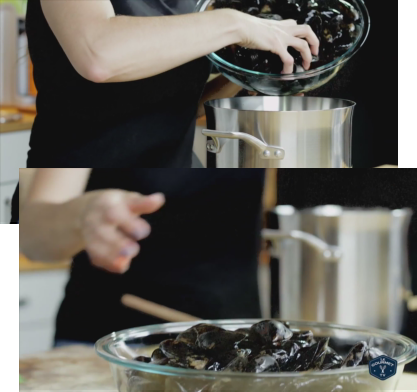}
         \end{subfigure}%
         \begin{subfigure}[t]{0.333\textwidth}
             \centering
             \includegraphics[width=\textwidth]{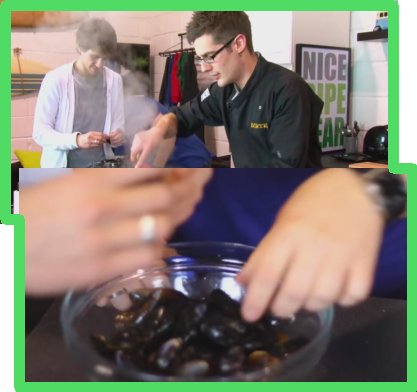}
         \end{subfigure}%
         \begin{subfigure}[t]{0.333\textwidth}
             \centering
             \includegraphics[width=\textwidth]{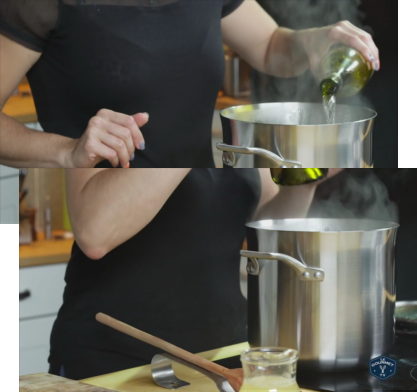}
         \end{subfigure}%
     \end{subfigure}%
     \hspace{0.019\textwidth}%
     \begin{subfigure}[t]{0.320\textwidth}
         \begin{subfigure}[t]{0.333\textwidth}
             \centering
            \includegraphics[width=\textwidth]{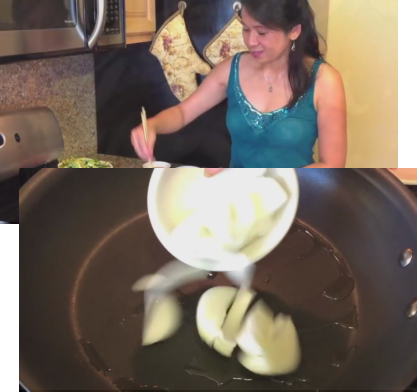}
         \end{subfigure}%
         \begin{subfigure}[t]{0.333\textwidth}
             \centering
             \includegraphics[width=\textwidth]{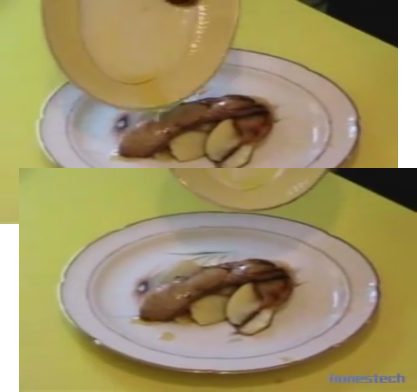}
         \end{subfigure}%
         \begin{subfigure}[t]{0.333\textwidth}
             \centering
             \includegraphics[width=\textwidth]{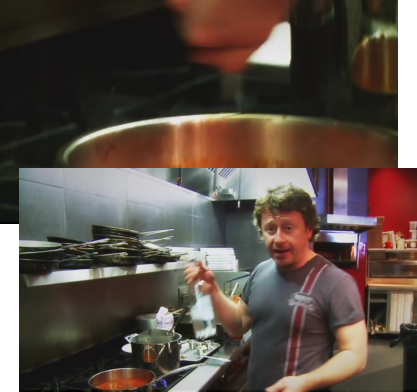}
         \end{subfigure}%
     \end{subfigure}%
     \hspace{0.019\textwidth}%
     \begin{subfigure}[t]{0.320\textwidth}
         \begin{subfigure}[t]{0.333\textwidth}
             \centering
            \includegraphics[width=\textwidth]{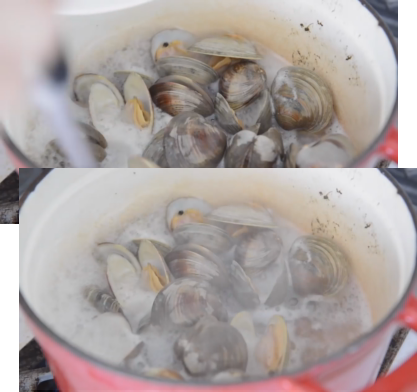}
         \end{subfigure}%
         \begin{subfigure}[t]{0.333\textwidth}
             \centering
             \includegraphics[width=\textwidth]{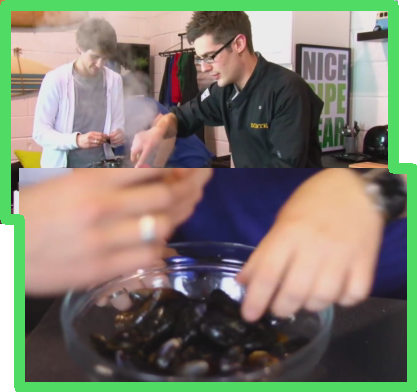}
         \end{subfigure}%
         \begin{subfigure}[t]{0.333\textwidth}
             \centering
             \includegraphics[width=\textwidth]{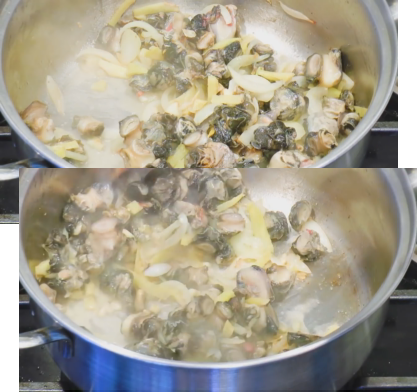}
         \end{subfigure}%
     \end{subfigure}
     
     \caption*{query: put the dish in the oven}
     \begin{subfigure}[t]{0.320\textwidth}
         \begin{subfigure}[t]{0.333\textwidth}
             \centering
            \includegraphics[width=\textwidth]{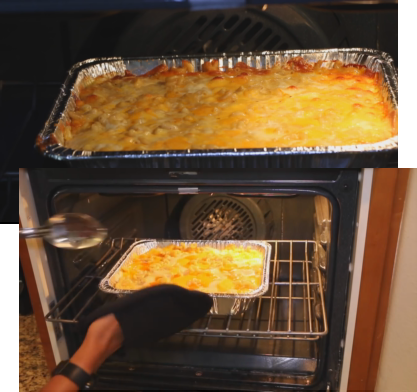}
         \end{subfigure}%
         \begin{subfigure}[t]{0.333\textwidth}
             \centering
             \includegraphics[width=\textwidth]{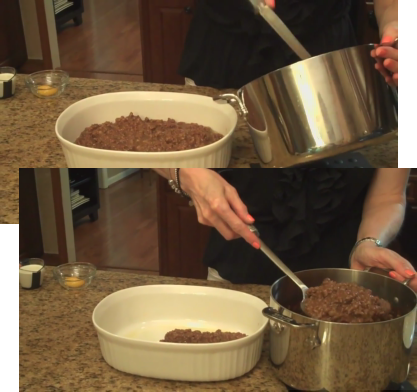}
         \end{subfigure}%
         \begin{subfigure}[t]{0.333\textwidth}
             \centering
             \includegraphics[width=\textwidth]{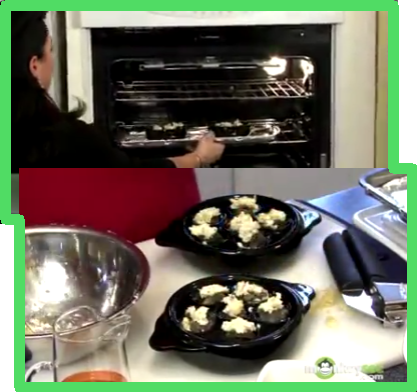}
         \end{subfigure}%
     \end{subfigure}%
     \hspace{0.019\textwidth}%
     \begin{subfigure}[t]{0.320\textwidth}
         \begin{subfigure}[t]{0.333\textwidth}
             \centering
            \includegraphics[width=\textwidth]{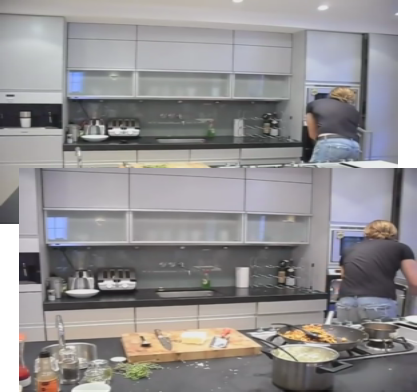}
         \end{subfigure}%
         \begin{subfigure}[t]{0.333\textwidth}
             \centering
             \includegraphics[width=\textwidth]{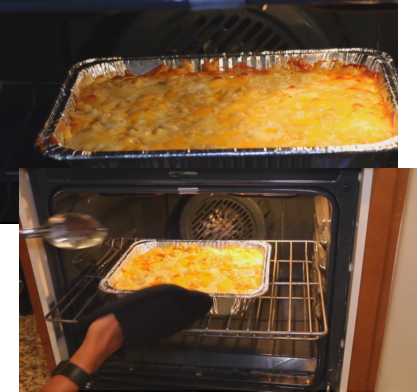}
         \end{subfigure}%
         \begin{subfigure}[t]{0.333\textwidth}
             \centering
             \includegraphics[width=\textwidth]{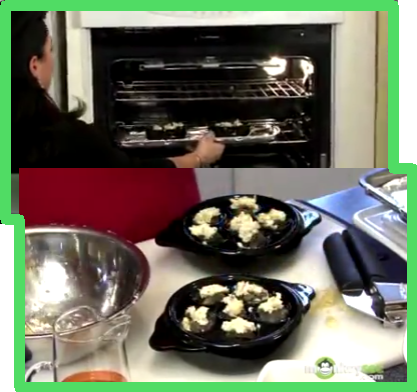}
         \end{subfigure}%
     \end{subfigure}%
     \hspace{0.019\textwidth}%
     \begin{subfigure}[t]{0.320\textwidth}
         \begin{subfigure}[t]{0.333\textwidth}
             \centering
            \includegraphics[width=\textwidth]{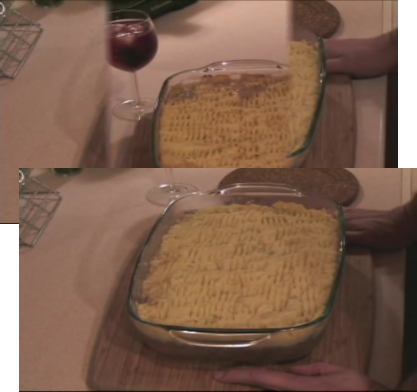}
         \end{subfigure}%
         \begin{subfigure}[t]{0.333\textwidth}
             \centering
             \includegraphics[width=\textwidth]{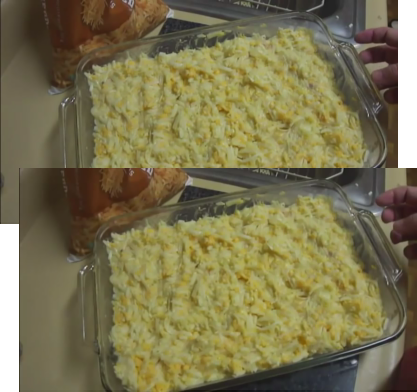}
         \end{subfigure}%
         \begin{subfigure}[t]{0.333\textwidth}
             \centering
             \includegraphics[width=\textwidth]{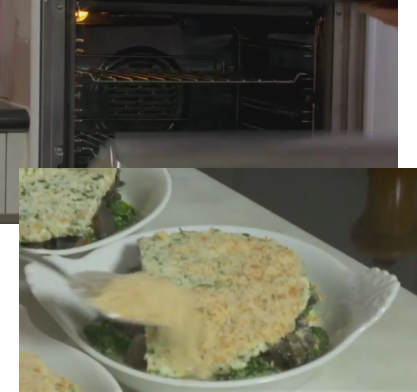}
         \end{subfigure}%
     \end{subfigure}
     
     \caption*{query: grill the ribs}
     \begin{subfigure}[t]{0.320\textwidth}
         \begin{subfigure}[t]{0.333\textwidth}
             \centering
            \includegraphics[width=\textwidth]{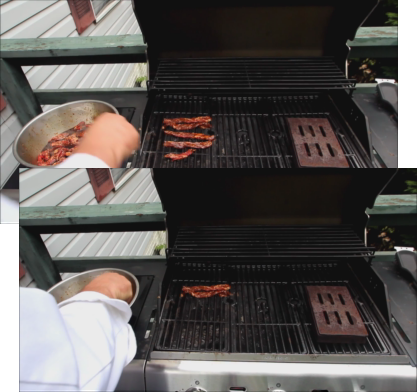}
         \end{subfigure}%
         \begin{subfigure}[t]{0.333\textwidth}
             \centering
             \includegraphics[width=\textwidth]{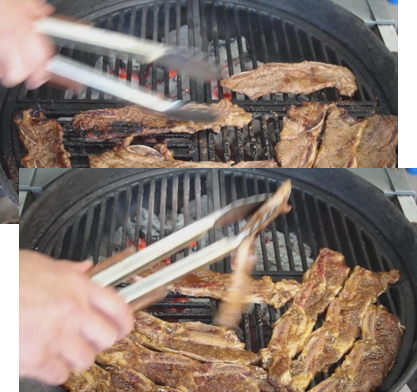}
         \end{subfigure}%
         \begin{subfigure}[t]{0.333\textwidth}
             \centering
             \includegraphics[width=\textwidth]{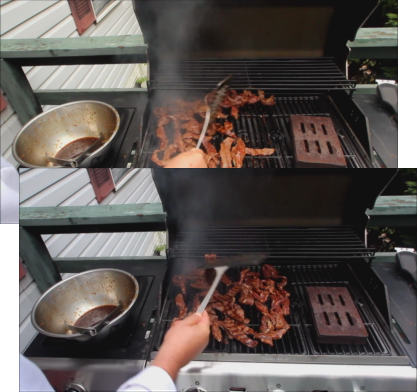}
         \end{subfigure}%
     \end{subfigure}%
     \hspace{0.019\textwidth}%
     \begin{subfigure}[t]{0.320\textwidth}
         \begin{subfigure}[t]{0.333\textwidth}
             \centering
            \includegraphics[width=\textwidth]{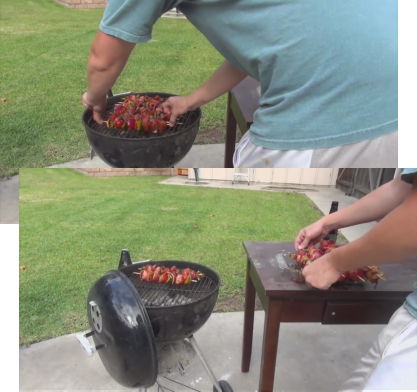}
         \end{subfigure}%
         \begin{subfigure}[t]{0.333\textwidth}
             \centering
             \includegraphics[width=\textwidth]{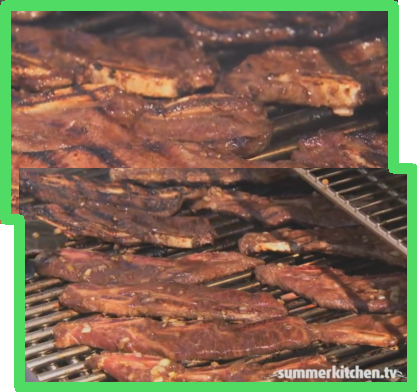}
         \end{subfigure}%
         \begin{subfigure}[t]{0.333\textwidth}
             \centering
             \includegraphics[width=\textwidth]{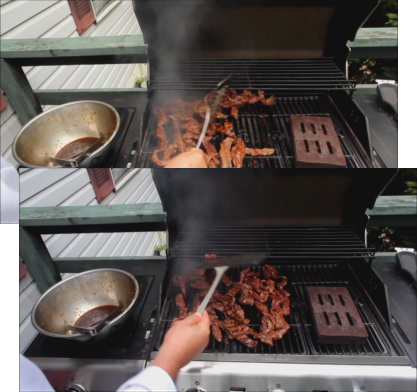}
         \end{subfigure}%
     \end{subfigure}%
     \hspace{0.019\textwidth}%
     \begin{subfigure}[t]{0.320\textwidth}
         \begin{subfigure}[t]{0.333\textwidth}
             \centering
            \includegraphics[width=\textwidth]{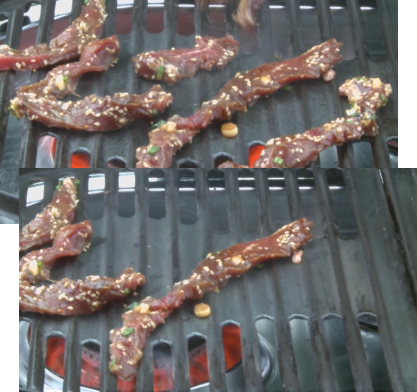}
         \end{subfigure}%
         \begin{subfigure}[t]{0.333\textwidth}
             \centering
             \includegraphics[width=\textwidth]{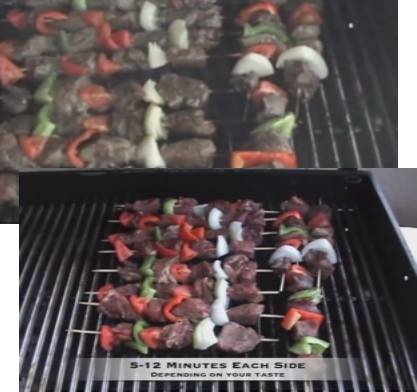}
         \end{subfigure}%
         \begin{subfigure}[t]{0.333\textwidth}
             \centering
             \includegraphics[width=\textwidth]{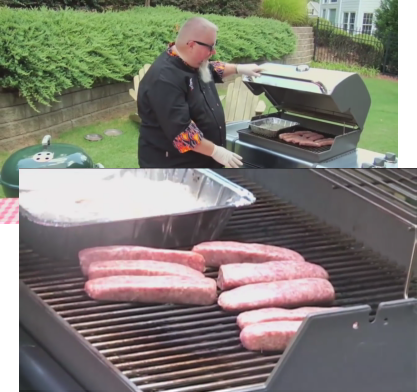}
         \end{subfigure}%
     \end{subfigure}
     
     \caption*{query: flip the pancake}
     \begin{subfigure}[t]{0.320\textwidth}
         \begin{subfigure}[t]{0.333\textwidth}
             \centering
            \includegraphics[width=\textwidth]{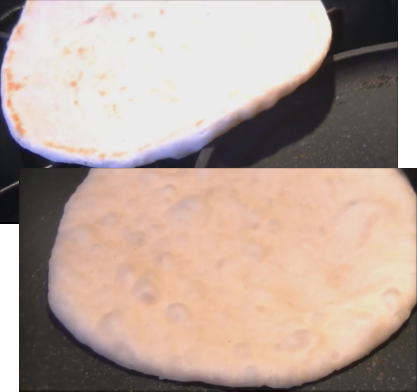}
         \end{subfigure}%
         \begin{subfigure}[t]{0.333\textwidth}
             \centering
             \includegraphics[width=\textwidth]{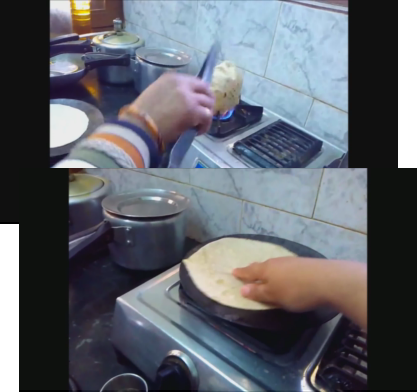}
         \end{subfigure}%
         \begin{subfigure}[t]{0.333\textwidth}
             \centering
             \includegraphics[width=\textwidth]{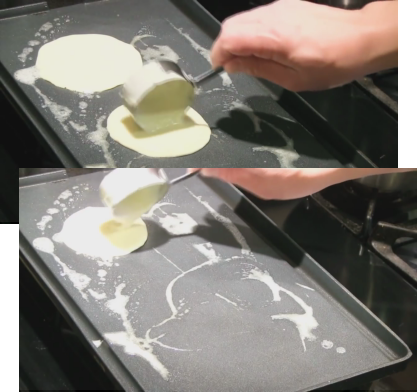}
         \end{subfigure}%
     \end{subfigure}%
     \hspace{0.019\textwidth}%
     \begin{subfigure}[t]{0.320\textwidth}
         \begin{subfigure}[t]{0.333\textwidth}
             \centering
            \includegraphics[width=\textwidth]{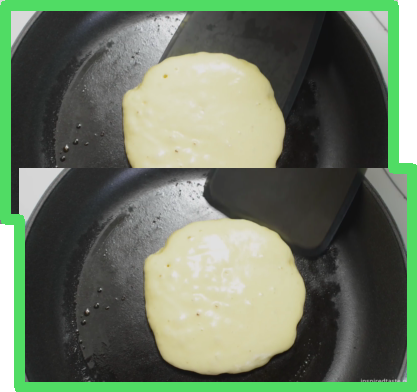}
         \end{subfigure}%
         \begin{subfigure}[t]{0.333\textwidth}
             \centering
             \includegraphics[width=\textwidth]{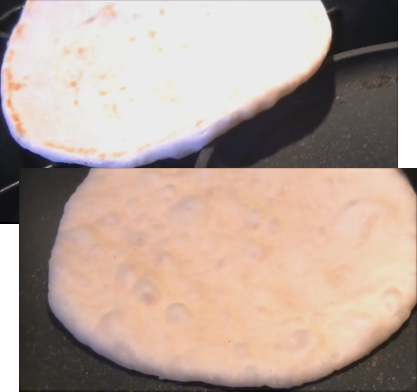}
         \end{subfigure}%
         \begin{subfigure}[t]{0.333\textwidth}
             \centering
             \includegraphics[width=\textwidth]{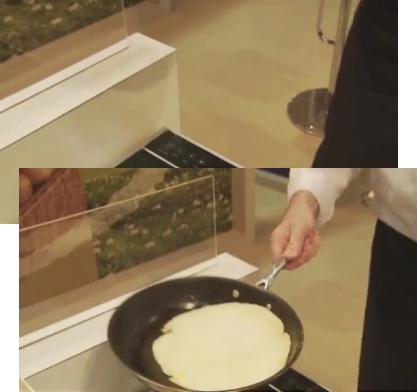}
         \end{subfigure}%
     \end{subfigure}%
     \hspace{0.019\textwidth}%
     \begin{subfigure}[t]{0.320\textwidth}
         \begin{subfigure}[t]{0.333\textwidth}
             \centering
            \includegraphics[width=\textwidth]{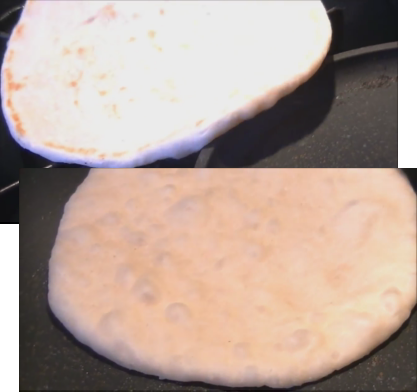}
         \end{subfigure}%
         \begin{subfigure}[t]{0.333\textwidth}
             \centering
             \includegraphics[width=\textwidth]{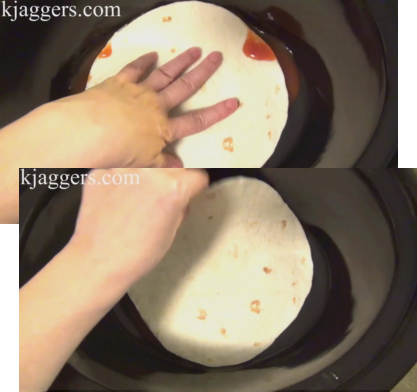}
         \end{subfigure}%
         \begin{subfigure}[t]{0.333\textwidth}
             \centering
             \includegraphics[width=\textwidth]{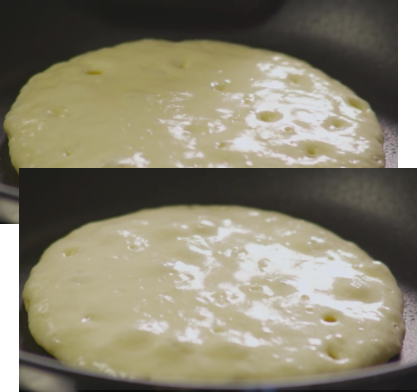}
         \end{subfigure}%
     \end{subfigure}

     \caption{More qualitative examples: top-3 zero-shot text-to-video retrieval results on the YouCook2 dataset for the proposed approach with self-attention based routing, the same one but without routing mechanism, and MIL-NCE* (* indicates that we used the same backbone as in our model). Correct video colored in green.}
    \label{fig:retrieval-supp}
\end{figure*}

\paragraph{Retrieval Results}
We present additional retrieval results for three models - our self-attention based routing method, our approach without routing, and MIL-NCE - in Figure \ref{fig:retrieval-supp}. Each column consists of the top-3 predictions for the given text query. Generally, routing achieves strong performance and retrieves visually varied videos; on the other hand, MIL-NCE tends to focus on specific objects or low-level visual cues leading to visually similar retrievals. The first three rows consists of examples where our self-attention based routing correctly retrieves videos but the other two methods do not. 
%Notably, our approach successfully handles the extremely specific query ``Put three rings of ketchup and two rings of mustard on the bottom bun". 
For general queries, like ``grill the ribs" and ``flip the pancakes" in the bottom two rows, there are many relevant videos to choose from. Only the no-routing method obtains the ``correct" video in its top-3 predictions, but these ``failure cases" for our method and MIL-NCE would be considered correct retrievals by human standards.

\begin{figure*}
    \centering
    
     \begin{subfigure}[t]{0.12\textwidth}
     \caption*{}
     \end{subfigure}%
     \hspace{0.02\textwidth}%
     \begin{subfigure}[t]{0.27\textwidth}
     \caption*{HowTo100M}
     \end{subfigure}%
     \hspace{0.02\textwidth}%
     \begin{subfigure}[t]{0.27\textwidth}
     \caption*{MSR-VTT}
     \end{subfigure}%
     \hspace{0.02\textwidth}%
     \begin{subfigure}[t]{0.27\textwidth}
     \caption*{YouCook2}
     \end{subfigure}%
     \vspace{-2mm}
    
    \begin{subfigure}[c]{0.12\textwidth}
            % \centering
            \stackinset{l}{0pt}{c}{0pt}{\Centerstack{
            \footnotesize \#49: pots/ \\
            \footnotesize bowls}}
            {\includegraphics[width=\textwidth]{Figures/text_to_video/blank.png}}
     \end{subfigure}%
     \hspace{0.02\textwidth}%
     \begin{subfigure}[t]{0.27 \textwidth}
        \begin{subfigure}[t]{0.49\textwidth}
            \centering
            \includegraphics[width=\textwidth]{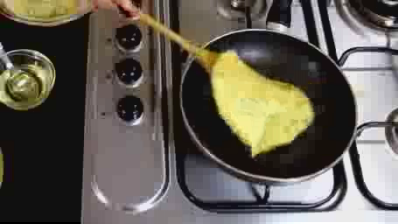}
         \end{subfigure}%
         \hspace{0.01\textwidth}%
         \begin{subfigure}[t]{0.49\textwidth}
             \centering
            \includegraphics[width=\textwidth]{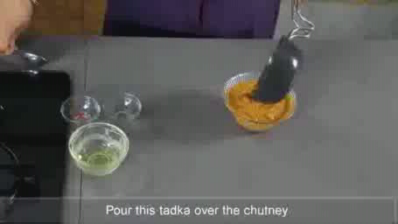}
         \end{subfigure}
         \begin{subfigure}[t]{0.49\textwidth}
            \centering
            \includegraphics[width=\textwidth]{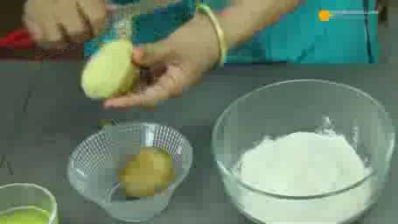}
         \end{subfigure}%
         \hspace{0.01\textwidth}%
         \begin{subfigure}[t]{0.49\textwidth}
             \centering
            \includegraphics[width=\textwidth]{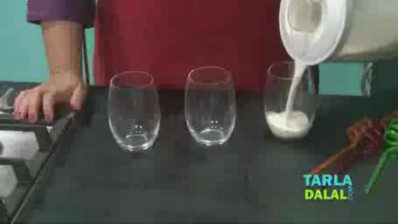}
         \end{subfigure}
     \end{subfigure}%
     \hspace{0.02\textwidth}%
     \begin{subfigure}[t]{0.27\textwidth}
        \begin{subfigure}[t]{0.49\textwidth}
            \centering
            \includegraphics[width=\textwidth]{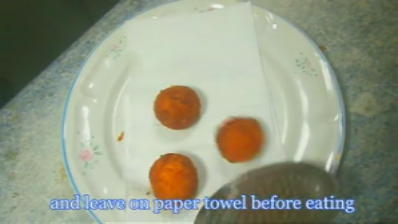}
         \end{subfigure}%
         \hspace{0.01\textwidth}%
         \begin{subfigure}[t]{0.49\textwidth}
             \centering
            \includegraphics[width=\textwidth]{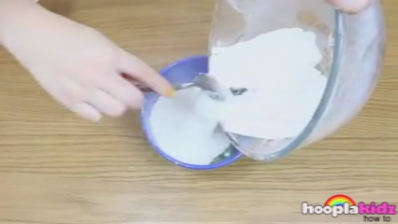}
         \end{subfigure}
         \begin{subfigure}[t]{0.49\textwidth}
            \centering
            \includegraphics[width=\textwidth]{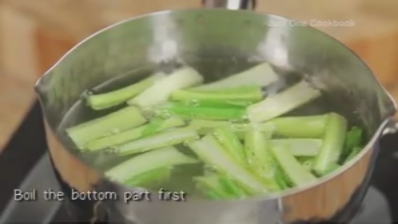}
         \end{subfigure}%
         \hspace{0.01\textwidth}%
         \begin{subfigure}[t]{0.49\textwidth}
             \centering
            \includegraphics[width=\textwidth]{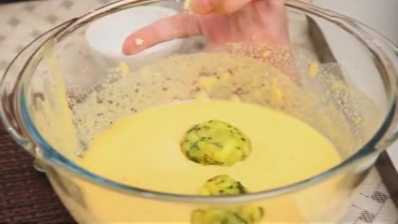}
         \end{subfigure}
     \end{subfigure}%
     \hspace{0.02\textwidth}%
     \begin{subfigure}[t]{0.27\textwidth}
        \begin{subfigure}[t]{0.49\textwidth}
            \centering
            \includegraphics[width=\textwidth]{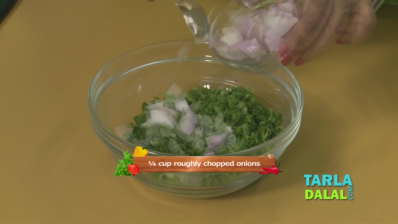}
         \end{subfigure}%
         \hspace{0.01\textwidth}%
         \begin{subfigure}[t]{0.49\textwidth}
             \centering
            \includegraphics[width=\textwidth]{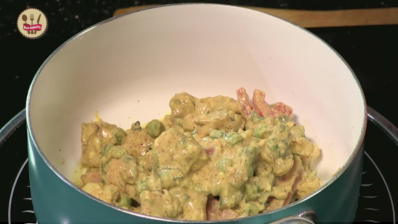}
         \end{subfigure}
         \begin{subfigure}[t]{0.49\textwidth}
            \centering
            \includegraphics[width=\textwidth]{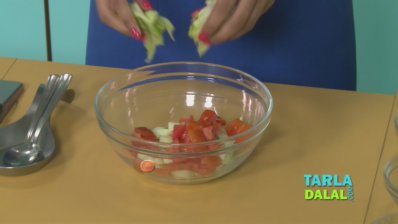}
         \end{subfigure}%
         \hspace{0.01\textwidth}%
         \begin{subfigure}[t]{0.49\textwidth}
             \centering
            \includegraphics[width=\textwidth]{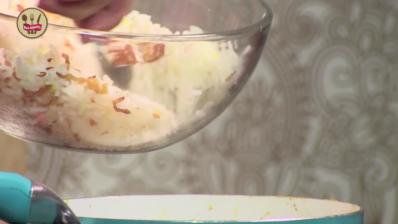}
         \end{subfigure}
     \end{subfigure}%
     \vspace{2mm}

    \begin{subfigure}[c]{0.12\textwidth}
            \centering
            \stackinset{l}{0pt}{c}{0pt}{\Centerstack{
            \footnotesize \#123: \\
            \footnotesize vegetables}}
            {\includegraphics[width=\textwidth]{Figures/text_to_video/blank.png}}
     \end{subfigure}%
     \hspace{0.02\textwidth}%
     \begin{subfigure}[t]{0.27 \textwidth}
        \begin{subfigure}[t]{0.49\textwidth}
            \centering
            \includegraphics[width=\textwidth]{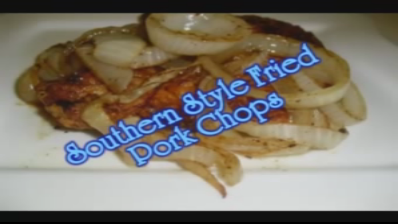}
         \end{subfigure}%
         \hspace{0.01\textwidth}%
         \begin{subfigure}[t]{0.49\textwidth}
             \centering
            \includegraphics[width=\textwidth]{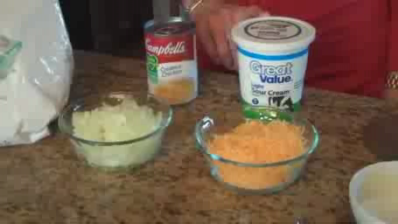}
         \end{subfigure}
         \begin{subfigure}[t]{0.49\textwidth}
            \centering
            \includegraphics[width=\textwidth]{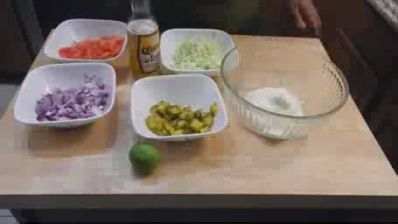}
         \end{subfigure}%
         \hspace{0.01\textwidth}%
         \begin{subfigure}[t]{0.49\textwidth}
             \centering
            \includegraphics[width=\textwidth]{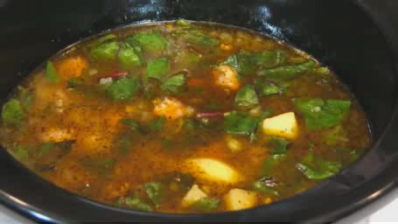}
         \end{subfigure}
     \end{subfigure}%
     \hspace{0.02\textwidth}%
     \begin{subfigure}[t]{0.27\textwidth}
        \begin{subfigure}[t]{0.49\textwidth}
            \centering
            \includegraphics[width=\textwidth]{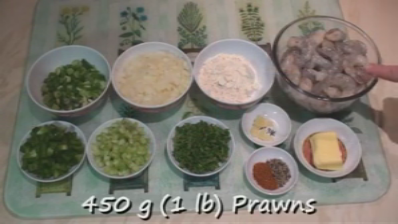}
         \end{subfigure}%
         \hspace{0.01\textwidth}%
         \begin{subfigure}[t]{0.49\textwidth}
             \centering
            \includegraphics[width=\textwidth]{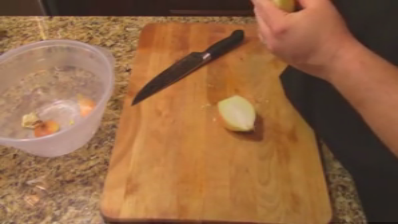}
         \end{subfigure}
         \begin{subfigure}[t]{0.49\textwidth}
            \centering
            \includegraphics[width=\textwidth]{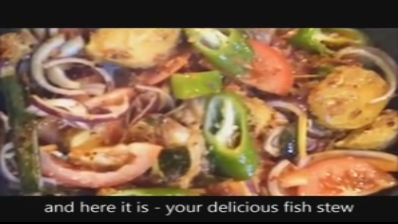}
         \end{subfigure}%
         \hspace{0.01\textwidth}%
         \begin{subfigure}[t]{0.49\textwidth}
             \centering
            \includegraphics[width=\textwidth]{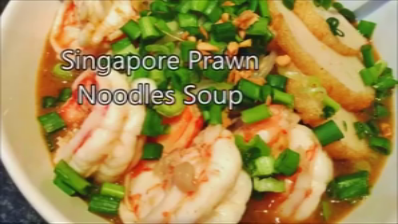}
         \end{subfigure}
     \end{subfigure}%
     \hspace{0.02\textwidth}%
     \begin{subfigure}[t]{0.27\textwidth}
        \begin{subfigure}[t]{0.49\textwidth}
            \centering
            \includegraphics[width=\textwidth]{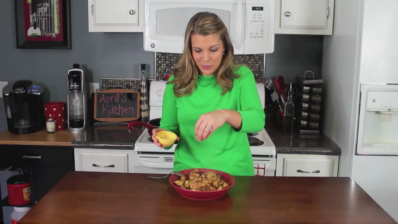}
         \end{subfigure}%
         \hspace{0.01\textwidth}%
         \begin{subfigure}[t]{0.49\textwidth}
             \centering
            \includegraphics[width=\textwidth]{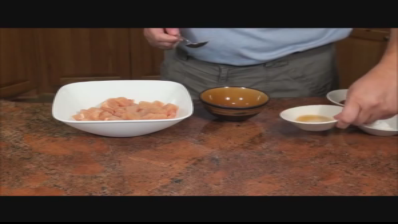}
         \end{subfigure}
         \begin{subfigure}[t]{0.49\textwidth}
            \centering
            \includegraphics[width=\textwidth]{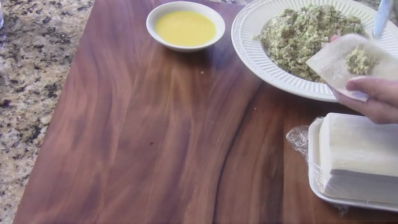}
         \end{subfigure}%
         \hspace{0.01\textwidth}%
         \begin{subfigure}[t]{0.49\textwidth}
             \centering
            \includegraphics[width=\textwidth]{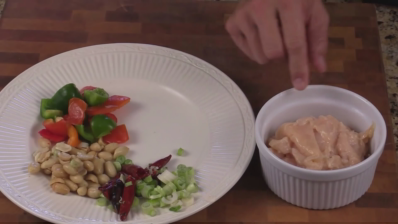}
         \end{subfigure}
     \end{subfigure}%
     \vspace{2mm}

     \begin{subfigure}[c]{0.12\textwidth}
            \centering
            \stackinset{l}{0pt}{c}{0pt}{\Centerstack{
            \footnotesize \#60: cooking
            }}
            {\includegraphics[width=\textwidth]{Figures/text_to_video/blank.png}}
     \end{subfigure}%
     \hspace{0.02\textwidth}%
     \begin{subfigure}[t]{0.27 \textwidth}
        \begin{subfigure}[t]{0.49\textwidth}
            \centering
            \includegraphics[width=\textwidth]{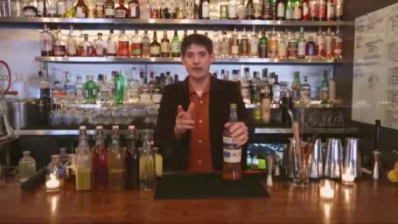}
         \end{subfigure}%
         \hspace{0.01\textwidth}%
         \begin{subfigure}[t]{0.49\textwidth}
             \centering
            \includegraphics[width=\textwidth]{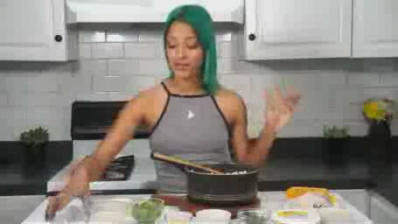}
         \end{subfigure}
         \begin{subfigure}[t]{0.49\textwidth}
            \centering
            \includegraphics[width=\textwidth]{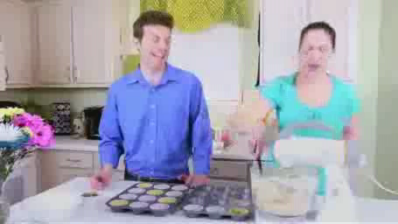}
         \end{subfigure}%
         \hspace{0.01\textwidth}%
         \begin{subfigure}[t]{0.49\textwidth}
             \centering
            \includegraphics[width=\textwidth]{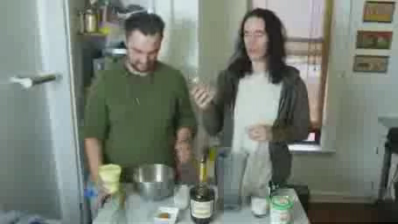}
         \end{subfigure}
     \end{subfigure}%
     \hspace{0.02\textwidth}%
     \begin{subfigure}[t]{0.27\textwidth}
        \begin{subfigure}[t]{0.49\textwidth}
            \centering
            \includegraphics[width=\textwidth]{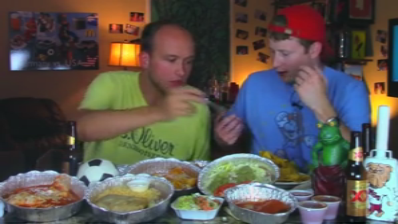}
         \end{subfigure}%
         \hspace{0.01\textwidth}%
         \begin{subfigure}[t]{0.49\textwidth}
             \centering
            \includegraphics[width=\textwidth]{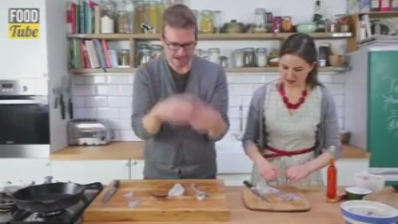}
         \end{subfigure}
         \begin{subfigure}[t]{0.49\textwidth}
            \centering
            \includegraphics[width=\textwidth]{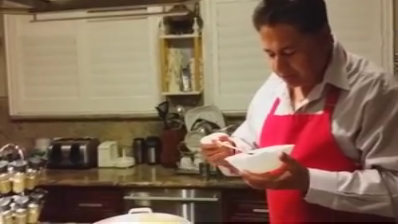}
         \end{subfigure}%
         \hspace{0.01\textwidth}%
         \begin{subfigure}[t]{0.49\textwidth}
             \centering
            \includegraphics[width=\textwidth]{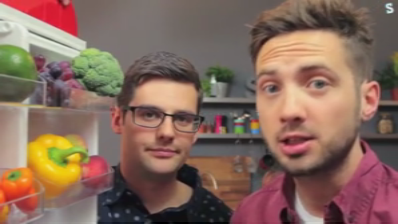}
         \end{subfigure}
     \end{subfigure}%
     \hspace{0.02\textwidth}%
     \begin{subfigure}[t]{0.27\textwidth}
        \begin{subfigure}[t]{0.49\textwidth}
            \centering
            \includegraphics[width=\textwidth]{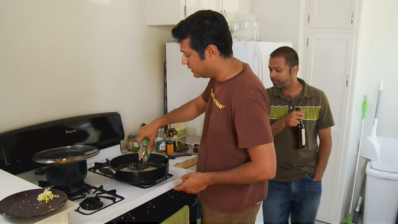}
         \end{subfigure}%
         \hspace{0.01\textwidth}%
         \begin{subfigure}[t]{0.49\textwidth}
             \centering
            \includegraphics[width=\textwidth]{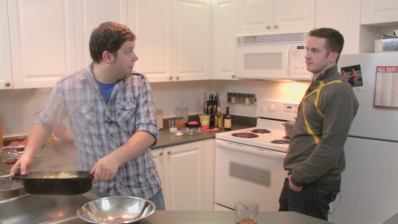}
         \end{subfigure}
         \begin{subfigure}[t]{0.49\textwidth}
            \centering
            \includegraphics[width=\textwidth]{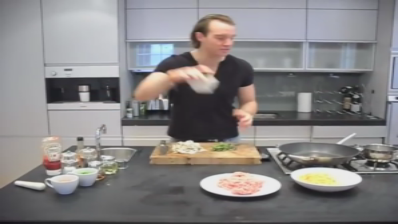}
         \end{subfigure}%
         \hspace{0.01\textwidth}%
         \begin{subfigure}[t]{0.49\textwidth}
             \centering
            \includegraphics[width=\textwidth]{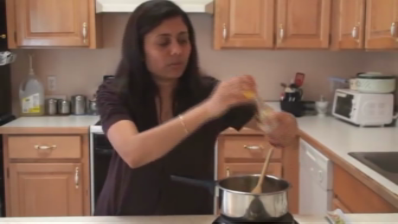}
         \end{subfigure}
     \end{subfigure}%
     \vspace{2mm}

     \begin{subfigure}[c]{0.12\textwidth}
            \stackinset{l}{0pt}{c}{0pt}{\Centerstack{
            \footnotesize \#70: hand \\
            \footnotesize make}}
            {\includegraphics[width=\textwidth]{Figures/text_to_video/blank.png}}
     \end{subfigure}%
     \hspace{0.02\textwidth}%
     \begin{subfigure}[t]{0.27 \textwidth}
        \begin{subfigure}[t]{0.49\textwidth}
            \centering
            \includegraphics[width=\textwidth]{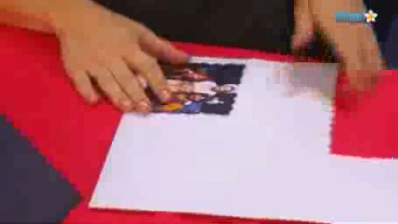}
         \end{subfigure}%
         \hspace{0.01\textwidth}%
         \begin{subfigure}[t]{0.49\textwidth}
             \centering
            \includegraphics[width=\textwidth]{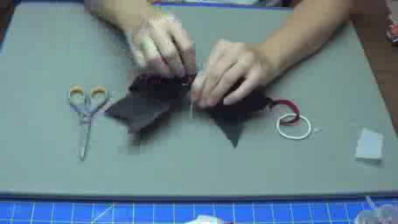}
         \end{subfigure}
         \begin{subfigure}[t]{0.49\textwidth}
            \centering
            \includegraphics[width=\textwidth]{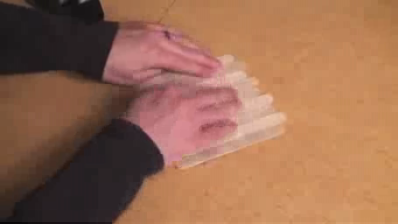}
         \end{subfigure}%
         \hspace{0.01\textwidth}%
         \begin{subfigure}[t]{0.49\textwidth}
             \centering
            \includegraphics[width=\textwidth]{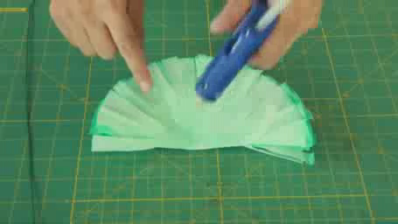}
         \end{subfigure}
     \end{subfigure}%
     \hspace{0.02\textwidth}%
     \begin{subfigure}[t]{0.27\textwidth}
        \begin{subfigure}[t]{0.49\textwidth}
            \centering
            \includegraphics[width=\textwidth]{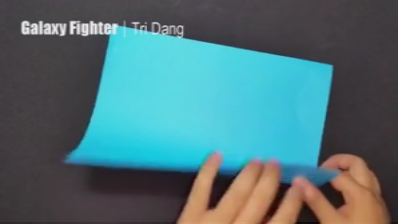}
         \end{subfigure}%
         \hspace{0.01\textwidth}%
         \begin{subfigure}[t]{0.49\textwidth}
             \centering
            \includegraphics[width=\textwidth]{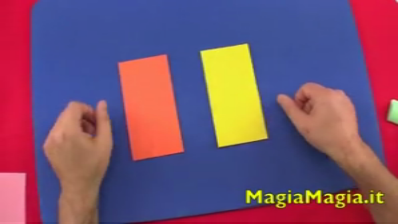}
         \end{subfigure}
         \begin{subfigure}[t]{0.49\textwidth}
            \centering
            \includegraphics[width=\textwidth]{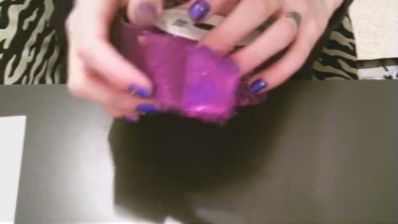}
         \end{subfigure}%
         \hspace{0.01\textwidth}%
         \begin{subfigure}[t]{0.49\textwidth}
             \centering
            \includegraphics[width=\textwidth]{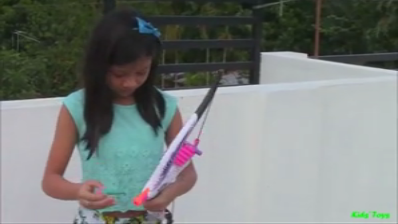}
         \end{subfigure}
     \end{subfigure}%
     \hspace{0.02\textwidth}%
     \begin{subfigure}[t]{0.27\textwidth}
        \begin{subfigure}[t]{0.49\textwidth}
            \centering
            \includegraphics[width=\textwidth]{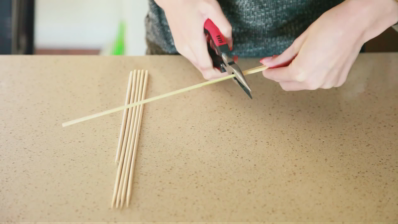}
         \end{subfigure}%
         \hspace{0.01\textwidth}%
         \begin{subfigure}[t]{0.49\textwidth}
             \centering
            \includegraphics[width=\textwidth]{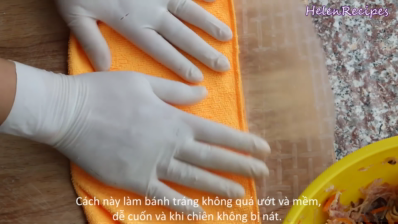}
         \end{subfigure}
         \begin{subfigure}[t]{0.49\textwidth}
            \centering
            \includegraphics[width=\textwidth]{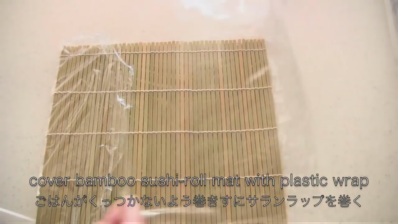}
         \end{subfigure}%
         \hspace{0.01\textwidth}%
         \begin{subfigure}[t]{0.49\textwidth}
             \centering
            \includegraphics[width=\textwidth]{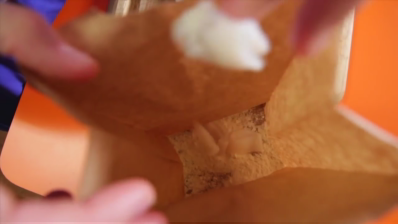}
         \end{subfigure}
     \end{subfigure}%
     \vspace{2mm}
     
     \begin{subfigure}[c]{0.12\textwidth}
            \centering
            \stackinset{l}{0pt}{c}{0pt}{\Centerstack{
            \footnotesize \#110: outdoor \\
            \footnotesize activity}}
            {\includegraphics[width=\textwidth]{Figures/text_to_video/blank.png}}
     \end{subfigure}%
     \hspace{0.02\textwidth}%
     \begin{subfigure}[t]{0.27 \textwidth}
        \begin{subfigure}[t]{0.49\textwidth}
            \centering
            \includegraphics[width=\textwidth]{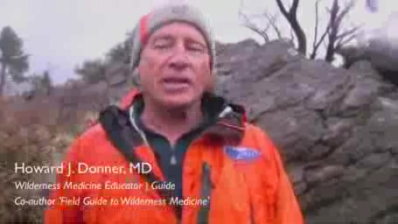}
         \end{subfigure}%
         \hspace{0.01\textwidth}%
         \begin{subfigure}[t]{0.49\textwidth}
             \centering
            \includegraphics[width=\textwidth]{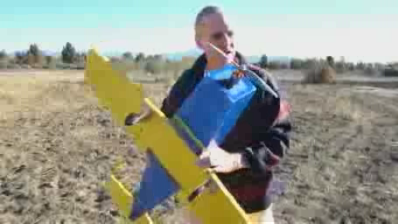}
         \end{subfigure}
         \begin{subfigure}[t]{0.49\textwidth}
            \centering
            \includegraphics[width=\textwidth]{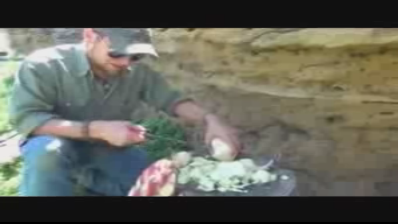}
         \end{subfigure}%
         \hspace{0.01\textwidth}%
         \begin{subfigure}[t]{0.49\textwidth}
             \centering
            \includegraphics[width=\textwidth]{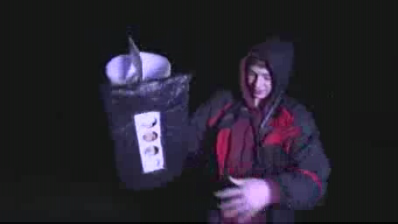}
         \end{subfigure}
     \end{subfigure}%
     \hspace{0.02\textwidth}%
     \begin{subfigure}[t]{0.27\textwidth}
        \begin{subfigure}[t]{0.49\textwidth}
            \centering
            \includegraphics[width=\textwidth]{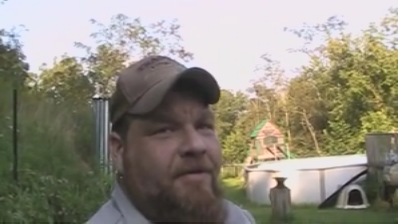}
         \end{subfigure}%
         \hspace{0.01\textwidth}%
         \begin{subfigure}[t]{0.49\textwidth}
             \centering
            \includegraphics[width=\textwidth]{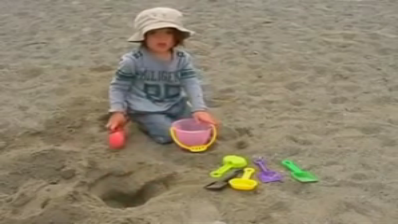}
         \end{subfigure}
         \begin{subfigure}[t]{0.49\textwidth}
            \centering
            \includegraphics[width=\textwidth]{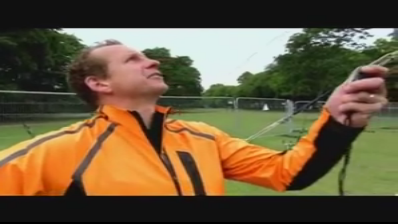}
         \end{subfigure}%
         \hspace{0.01\textwidth}%
         \begin{subfigure}[t]{0.49\textwidth}
             \centering
            \includegraphics[width=\textwidth]{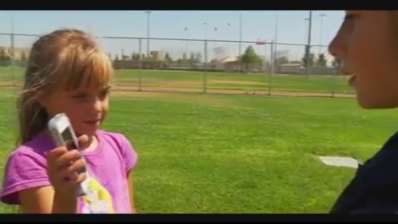}
         \end{subfigure}
     \end{subfigure}%
     \hspace{0.02\textwidth}%
     \begin{subfigure}[t]{0.27\textwidth}
        \begin{subfigure}[t]{0.49\textwidth}
            \centering
            \includegraphics[width=\textwidth]{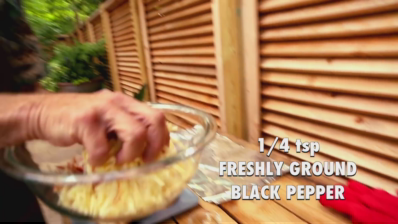}
         \end{subfigure}%
         \hspace{0.01\textwidth}%
         \begin{subfigure}[t]{0.49\textwidth}
             \centering
            \includegraphics[width=\textwidth]{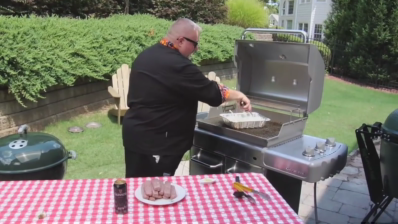}
         \end{subfigure}
         \begin{subfigure}[t]{0.49\textwidth}
            \centering
            \includegraphics[width=\textwidth]{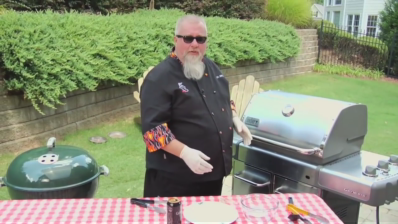}
         \end{subfigure}%
         \hspace{0.01\textwidth}%
         \begin{subfigure}[t]{0.49\textwidth}
             \centering
            \includegraphics[width=\textwidth]{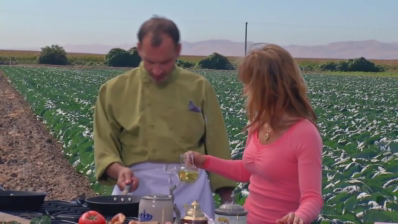}
         \end{subfigure}
     \end{subfigure}%
     \vspace{2mm}
     
     \begin{subfigure}[c]{0.12\textwidth}
            \centering
            \stackinset{l}{0pt}{c}{0pt}{\Centerstack{
            \footnotesize \#85: repair}}
            {\includegraphics[width=\textwidth]{Figures/text_to_video/blank.png}}
     \end{subfigure}%
     \hspace{0.02\textwidth}%
     \begin{subfigure}[t]{0.27 \textwidth}
        \begin{subfigure}[t]{0.49\textwidth}
            \centering
            \includegraphics[width=\textwidth]{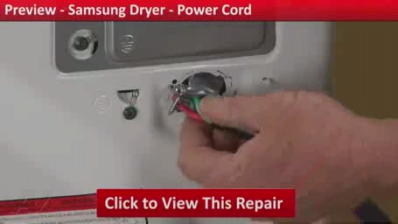}
         \end{subfigure}%
         \hspace{0.01\textwidth}%
         \begin{subfigure}[t]{0.49\textwidth}
             \centering
            \includegraphics[width=\textwidth]{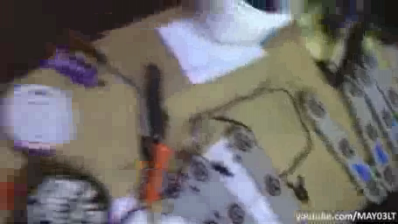}
         \end{subfigure}
         \begin{subfigure}[t]{0.49\textwidth}
            \centering
            \includegraphics[width=\textwidth]{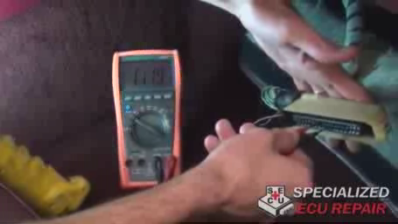}
         \end{subfigure}%
         \hspace{0.01\textwidth}%
         \begin{subfigure}[t]{0.49\textwidth}
             \centering
            \includegraphics[width=\textwidth]{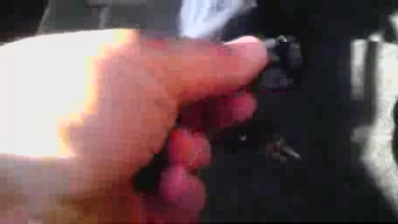}
         \end{subfigure}
     \end{subfigure}%
     \hspace{0.02\textwidth}%
     \begin{subfigure}[t]{0.27\textwidth}
        \begin{subfigure}[t]{0.49\textwidth}
            \centering
            \includegraphics[width=\textwidth]{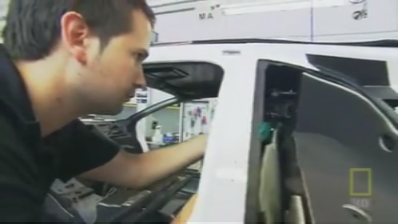}
         \end{subfigure}%
         \hspace{0.01\textwidth}%
         \begin{subfigure}[t]{0.49\textwidth}
             \centering
            \includegraphics[width=\textwidth]{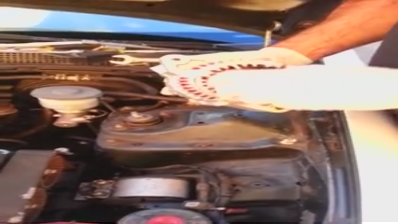}
         \end{subfigure}
         \begin{subfigure}[t]{0.49\textwidth}
            \centering
            \includegraphics[width=\textwidth]{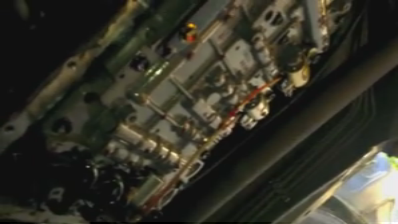}
         \end{subfigure}%
         \hspace{0.01\textwidth}%
         \begin{subfigure}[t]{0.49\textwidth}
             \centering
            \includegraphics[width=\textwidth]{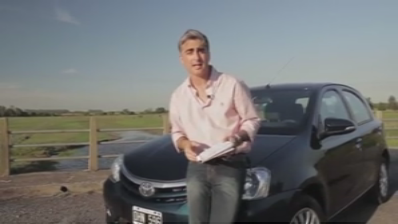}
         \end{subfigure}
     \end{subfigure}%
     \hspace{0.02\textwidth}%
     \begin{subfigure}[t]{0.27\textwidth}
        \begin{subfigure}[t]{0.49\textwidth}
            \centering
            \includegraphics[width=\textwidth]{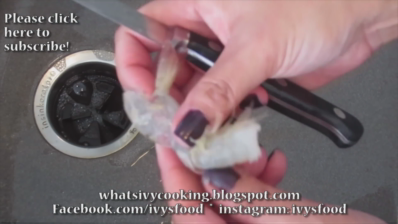}
         \end{subfigure}%
         \hspace{0.01\textwidth}%
         \begin{subfigure}[t]{0.49\textwidth}
             \centering
            \includegraphics[width=\textwidth]{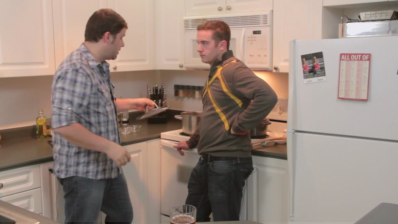}
         \end{subfigure}
         \begin{subfigure}[t]{0.49\textwidth}
            \centering
            \includegraphics[width=\textwidth]{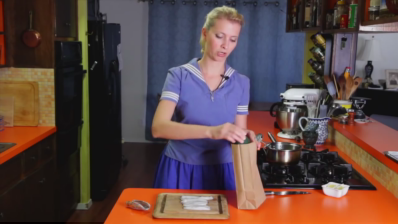}
         \end{subfigure}%
         \hspace{0.01\textwidth}%
         \begin{subfigure}[t]{0.49\textwidth}
             \centering
            \includegraphics[width=\textwidth]{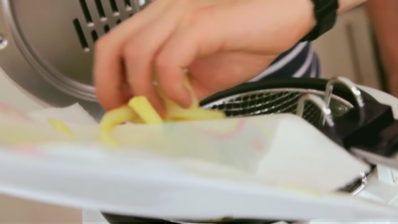}
         \end{subfigure}
     \end{subfigure}%
     \vspace{2mm}

     \begin{subfigure}[c]{0.12\textwidth}
            \centering
            \stackinset{l}{0pt}{c}{0pt}{\Centerstack{
            \footnotesize \#28: pets/ \\
            \footnotesize animals
            }}
            {\includegraphics[width=\textwidth]{Figures/text_to_video/blank.png}}
     \end{subfigure}%
     \hspace{0.02\textwidth}%
     \begin{subfigure}[t]{0.27 \textwidth}
        \begin{subfigure}[t]{0.49\textwidth}
            \centering
            \includegraphics[width=\textwidth]{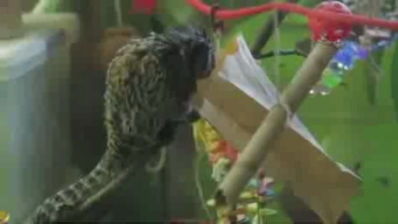}
         \end{subfigure}%
         \hspace{0.01\textwidth}%
         \begin{subfigure}[t]{0.49\textwidth}
             \centering
            \includegraphics[width=\textwidth]{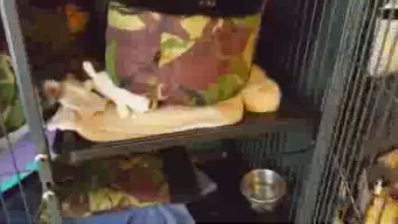}
         \end{subfigure}
         \begin{subfigure}[t]{0.49\textwidth}
            \centering
            \includegraphics[width=\textwidth]{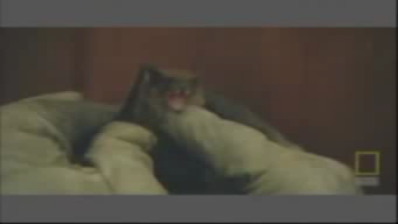}
         \end{subfigure}%
         \hspace{0.01\textwidth}%
         \begin{subfigure}[t]{0.49\textwidth}
             \centering
            \includegraphics[width=\textwidth]{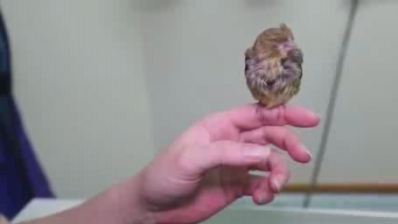}
         \end{subfigure}
     \end{subfigure}%
     \hspace{0.02\textwidth}%
     \begin{subfigure}[t]{0.27\textwidth}
        \begin{subfigure}[t]{0.49\textwidth}
            \centering
            \includegraphics[width=\textwidth]{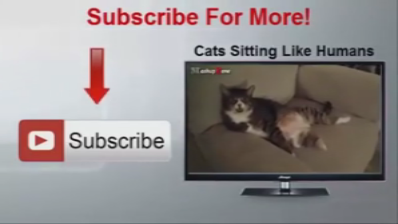}
         \end{subfigure}%
         \hspace{0.01\textwidth}%
         \begin{subfigure}[t]{0.49\textwidth}
             \centering
            \includegraphics[width=\textwidth]{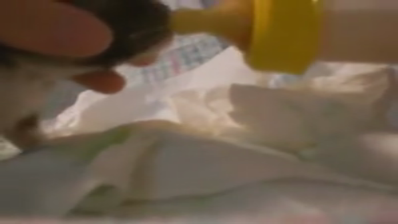}
         \end{subfigure}
         \begin{subfigure}[t]{0.49\textwidth}
            \centering
            \includegraphics[width=\textwidth]{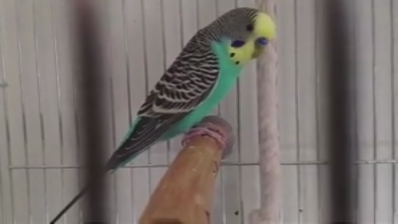}
         \end{subfigure}%
         \hspace{0.01\textwidth}%
         \begin{subfigure}[t]{0.49\textwidth}
             \centering
            \includegraphics[width=\textwidth]{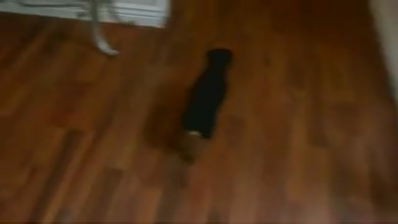}
         \end{subfigure}
     \end{subfigure}%
     \hspace{0.02\textwidth}%
     \begin{subfigure}[t]{0.27\textwidth}
        \begin{subfigure}[t]{0.49\textwidth}
            \centering
            \includegraphics[width=\textwidth]{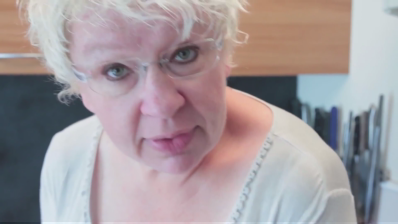}
         \end{subfigure}%
         \hspace{0.01\textwidth}%
         \begin{subfigure}[t]{0.49\textwidth}
             \centering
            \includegraphics[width=\textwidth]{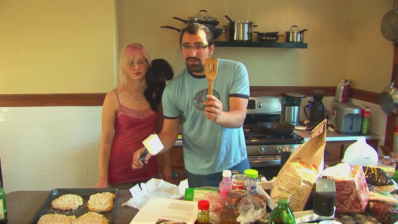}
         \end{subfigure}
         \begin{subfigure}[t]{0.49\textwidth}
            \centering
            \includegraphics[width=\textwidth]{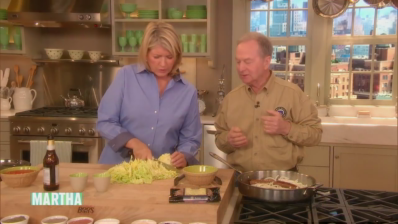}
         \end{subfigure}%
         \hspace{0.01\textwidth}%
         \begin{subfigure}[t]{0.49\textwidth}
             \centering
            \includegraphics[width=\textwidth]{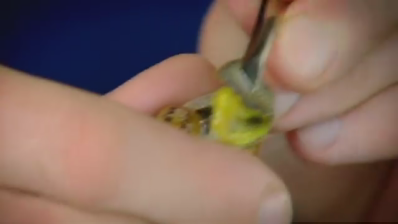}
         \end{subfigure}
     \end{subfigure}%
     \vspace{2mm}
     
     \caption{Extended figure with examples of capsule highest activations: top-4 videos with the highest activation for the particular capsule for the HowTo100M, MSR-VTT, and YouCook2 datasets. Labels: \#number of capsule: assumed learned ``concept". }
    \label{fig:capsule-acts-supp}
\end{figure*}

\paragraph{Capsule Activations}
In Figure \ref{fig:capsule-acts-supp}, we include videos which correspond to various capsules' highest activations across three different datasets: HowTo100M, MSR-VTT, and YouCook2. The concepts tend to remain consistent across the different datasets. In the final three rows, we present examples where the concept does not exist within the target dataset: ``repair", ``games", and ``pets/animals" are not present within the cooking dataset, YouCook2. Since these capsules learn to represent these specific concepts/entities, their highest activation corresponds to seemingly random videos.

\end{document}